\documentclass[10pt,journal,compsoc]{IEEEtran}
\pdfoutput=1
\usepackage{hyperref}
\usepackage{url}
\usepackage{soul}
\usepackage{subcaption}
\usepackage{enumitem}

\usepackage[monochrome]{color}
\usepackage{tikz}
\usepackage{wrapfig}

\usepackage{cite}
\usepackage{graphics}
\usepackage{graphicx}
\usepackage{amsmath}

\begin{document}
%
\title{Exploiting Activation based Gradient Output Sparsity to Accelerate Backpropagation in CNNs}
%
%
%
%

\author{Anup~Sarma,
        Sonali~Singh,
        Huaiapn~Jiang, 
        Ashutosh Pattnaik,
        Asit K. Mishra, 
        Mahmut Taylan Kandemir,
        Vijaykrishnan Narayanan
        and~Chita~R~Das\\
        The Pennsylvania State University
}

\markboth{Journal of \LaTeX\ Class Files,~Vol.~14, No.~8, August~2015}%
{Shell \MakeLowercase{\textit{et al.}}: Bare Demo of IEEEtran.cls for Computer Society Journals}

\IEEEtitleabstractindextext{%
\begin{abstract}
Machine/deep-learning (ML/DL) based techniques are emerging as a driving force behind many cutting-edge technologies, achieving high accuracy on computer vision workloads such as image classification and object detection. However, training these models involving large parameters is both time-consuming and energy-hogging. In this regard, several prior works have advocated for sparsity to speed up the of DL training and more so, the inference phase.
This work begins with the observation that during training, sparsity in the forward and backward passes are correlated. In that context, we investigate two types of sparsity (input and output type) inherent in gradient descent-based optimization algorithms and propose a hardware micro-architecture to leverage the same. Our experimental results use five state-of-the-art CNN models on the Imagenet dataset, and show back propagation speedups in the range of 1.69$\times$ to 5.43$\times$, compared to the dense baseline execution. By exploiting sparsity in both the forward and backward passes, speedup improvements range from 1.68$\times$ to 3.30$\times$ over the sparsity-agnostic baseline execution.  Our work also achieves significant reduction in training iteration time over several previously proposed dense as well as sparse accelerator based platforms, in addition to achieving order of magnitude energy efficiency improvements over GPU based execution.


\end{abstract}

}

\maketitle

\thispagestyle{plain}
\pagestyle{plain}








\section{Introduction}

Deep learning-based convolutional neural networks (CNNs) have outperformed traditional rule-based algorithms and have achieved state-of-the-art accuracy on vision workloads such as image classification~\cite{russakovsky2015imagenet,ciregan2012multi}, object detection~\cite{fasterRCNN}, and semantic segmentation~\cite{long2015fully,paszke2016enet}. As these networks grow deeper, wider and more sophisticated, training such deep CNN-based models requires significant compute and energy budget. Therefore, it is imperative to address the ever-increasing computational demands of training such networks.

\begin{figure}
   \centering
   \includegraphics[width=0.60\linewidth]{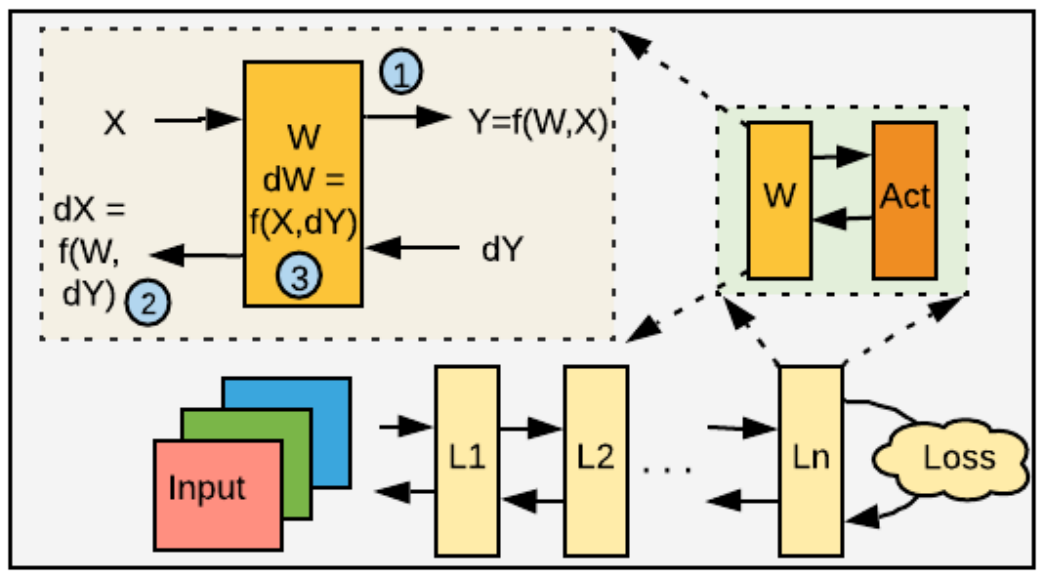}
   \caption{ Within a training step, an input feature map (X) is processed by a layer to generate the output feature map (Y) in forward pass, and the input gradient map (dY) is processed by the layer to produce the output gradient map (dX) in backward pass, along with weight gradient (dW).}
   \label{fig:Intro_figure}
\end{figure}

A typical vision based CNN architecture, as depicted in Fig.~\ref{fig:Intro_figure}, can be abstracted as a connected set of $n$ layers $L1, L2,.., Ln$, where \textit{most} of the layers are composed of a weight layer followed by an activation layer. Here, a convolution (CONV) acts as the weight layer (W),
while a Rectified Linear Unit (ReLU) is widely used as an activation function~\cite{VGGNet, GoogleNet, ResNet, DenseNet, howard2017mobilenets}. A representative training step is shown in Fig.~\ref{fig:Intro_figure}, which transforms an input feature map (X) in the forward pass (FP) and gradient values (dY) in the backward pass (BP) using the weight parameters (W), and also computes the weight gradients (dW) using the loss function during Weight Gradient(WG) phase.
Generalized matrix multiplication (GEMM) is a key building block of such transformations, and offers opportunities to take advantage of regular compute and communication characteristics of GEMM operation. Towards this end, several dedicated accelerator designs have been proposed~\cite{chen2014diannao, chen2014dadiannao, du2015shidiannao, liu2015pudiannao, chen2017eyeriss}, including Tensor Processing Units~\cite{TPU} and fixed function accelerator for GPUs~\cite{volta}.
Such accelerators mainly target dense GEMM kernels in which both the inputs to the GEMM kernel (weight values and neuron activation values in a layer) are dense matrices and the output (activation for the next layer) is also a dense matrix.

However, an important characteristic of all performant CNN models is that they exhibit significant ``sparsity" in various data structures: sparsity in feature map activations as well as in weights~\cite{wen2016learning,kim2016evaluation}. 
Sparsity of neuron activation is a well known phenomenon, which is present across all the layers of ReLU-based CNNs~\cite{relucnn}. Sparsity of weights, on the other hand, is achieved by pruning weight values, which are close to zero and is usually done in a post-training phase as a fine-tuning step targeted for model compression \cite{han2015learning,polino2018model}.
Several hardware architectures have been proposed in literature to take advantage of weight and activation sparsity~\cite {han2016eie, bell2008efficient, vazquez2010improving, liu2013efficient} in the forward pass (during inference).
These proposals exploit sparsity to reduce the number of compute operations providing opportunities for improved performance and lower energy consumption. 
As shown in Fig.~\ref{fig:Intro_figure}, $Y = f(W, X)$ computation can be performed more efficiently by considering sparsity in $X$.

\begin{figure}[t]
   \centering
   \includegraphics[width=0.90\linewidth]{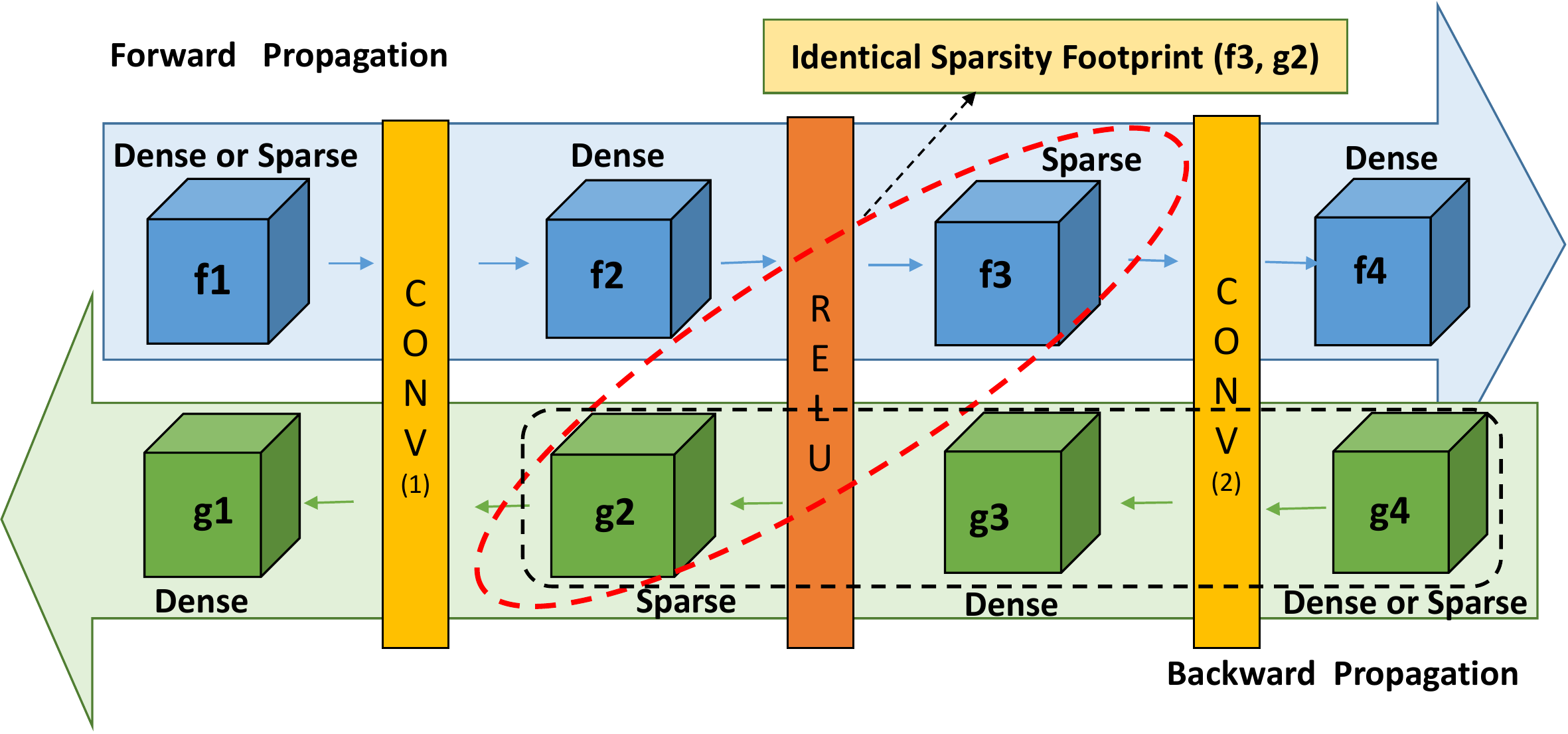}
   \caption{Forward propagation of feature maps and Backward propagation of error gradients through different layers during a neural network training step.}
   \label{fig:FP_BP}
\end{figure}

Traditionally, CNNs have CONV-ReLU-CONV chains (shown in Fig.~\ref{fig:FP_BP}), where the ReLU layer is the activation layer. The input to the  CONV(1) layer \textbf{f1} can be sparse or dense,  depending on previous layer type. However,  \textbf{f2}, which is the output of CONV(1), is usually dense. When this \textbf{f2} is passed through the ReLU layer (computes $max(0,x)$), the output \textbf{f3} will consist of zero values that correspond to the negative values of \textbf{f2}, and thus, it is sparse in nature. When computing \textbf{f4} using \textbf{f3} as an input, we can leverage sparsity of \textbf{f3} to skip the zero-valued computation. We refer to this type of sparsity as \textit{input sparsity}, and most of the prior works deal with exploiting such input sparsity.

It has also been observed that neuron gradients ($dY$) can be sparse, and therefore, there exists opportunity for utilizing input sparsity during output gradient computation. Thus, several prior works have also sought to exploit gradient sparsity in the backward pass of training \cite{spartann, sparten,sparsetrain, rajbhandari2017optimizing}. However, most of these prior works only rely on input sparsity technique, which limit their ability to exploit gradient sparsity in the presence of batch normalization layer. Our goal, in this work, is to leverage sparsity in the backward pass of training during neuron gradient computation, not only by exploiting the \textbf{input} sparsity, but also the \textbf{output} sparsity of error gradients. This is achieved by leveraging \textit{apriori} information which neuron output locations are \textit{going to be zero}, further illustrated in Fig.2.




Thus, during back-propagation stage of error gradients, \textbf{g4} can be dense or sparse depending on the subsequent layer. Now, \textbf{g4} is used to compute \textbf{g3}, and this output is usually dense. When \textbf{g3} is passed through the ReLU layer, we find that it generates a sparse output \textbf{g2} with an \textit{identical sparsity footprint} as \textbf{f3}.
In other words, sparsity at \textbf{g2} can be known irrespective of its input \textbf{g3} (detailed theoretical analysis in Section \ref{identical_sparsity}). 
Because the sparsity footprint of \textbf{f3} is already known during forward pass, we can leverage this information to avoid any computation corresponding to those output locations at \textbf{g3}. 
We refer to this type of sparsity as \textit{output sparsity} in the backward pass, and no prior work has identified scope for exploiting this opportunity as the high dimensional of tensors involved in CNNs make it extremely difficult to identify such patterns. Note that if \textbf{g4} was sparse to begin with, we could also leverage input sparsity of \textbf{g4}, in addition to exploiting output sparsity at \textbf{g3}, as determined by sparsity at \textbf{f3}. This can lead to further gains in both performance and energy efficiency.

After identifying opportunities to leverage sparsity in the forward pass (\textit{input}) and backward pass (\textit{output} and/or \textit{input}), we propose a novel micro-architecture that is able to skip computations based on sparsity type. 
The design incorporates several principles to efficiently perform computations: input double buffering to reduce lane level stall cycles, weight blocking to achieve better memory bandwidth utilization and a re-configurable adder tree to enhance utilization of compute building blocks. 
To account for the asymmetry of sparsity distribution, we also propose a work redistribution mechanism that efficiently handles load imbalance at runtime.

Towards this end, the \textbf{primary contributions} of this paper are the following:

$\bullet$ To the best of our knowledge, this is the first work that identifies the \textit{symmetry} of sparsity relationship between the forward and backward passes and \textit{effectively} exploits them in the form of \textit{output sparsity} in the backward pass, in addition to \textit{input} sparsity which can be leveraged at both forward and backward passes. We provide qualitative reasoning for the sources of such sparsity and also mathematically formulate a relationship between activation sparsity and neuron gradients.


$\bullet$ Based on the insights from the theoretical analysis, we propose a novel hardware architecture to exploit both types of sparsities. We provide the design principles of the proposed architecture and incorporate necessary hardware mechanisms to facilitate computation skipping. 

$\bullet$ We evaluate five state-of-the-art CNN models (VGGNet \cite{VGGNet}, ResNet18 \cite{ResNet}, GoogLeNet~\cite{GoogleNet}, DenseNet~\cite{DenseNet} and MobileNet~\cite{howard2017mobilenets}) to demonstrate the benefits of our approach. By exploiting sparsity in both the forward and backward passes, speedup improvements range from 1.68$\times$ to 3.30$\times$ over the sparsity-agnostic baseline execution. In addition, we also show performance and energy-efficiency benefits over several previously proposed dense as well as sparse accelerator platforms, while also achieving order of magnitude energy efficiency over GPU based execution.

\section{Motivation and Related Work} \label{sec:bgmv}

 \begin{figure*}
   \centering
   \includegraphics[width=1.0\linewidth]{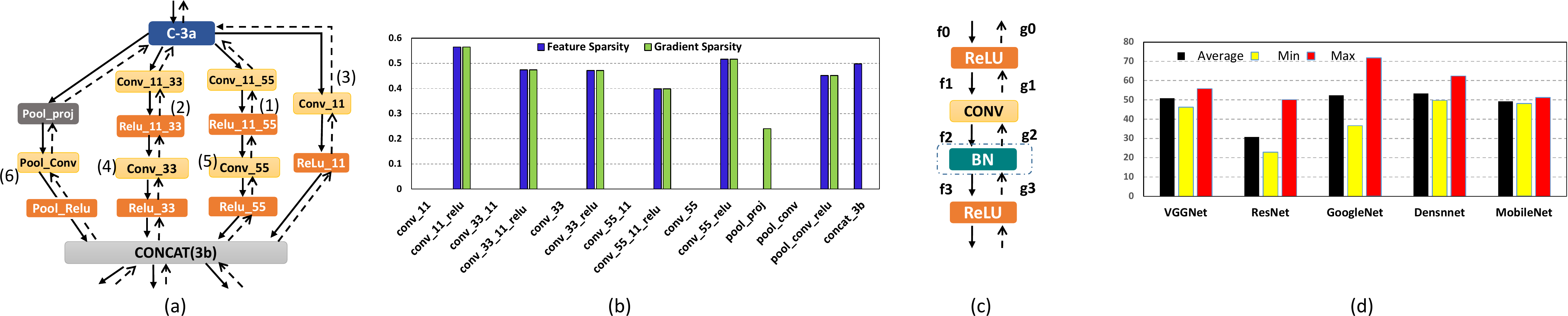}
   \caption{(a) Inception \textit{(3b block)} from GoogLeNet. Solid arrows represent forward propagation, while broken arrows represent backward propagation (b) Fraction of sparsity for feature and gradient at the output of different layers of Inception-3b block (c) Network with Batch Normalization layer which prevents input sparsity in backward pass (d) Fraction of average minimum, average maximum and total average sparsity across a batch size of 16 for a randomly chosen training step}
   \label{fig:Motivation_Fig}
\end{figure*}

This sections summarizes the motivation for our work followed by a brief overview  of related work that explores sparsity in CNNs. 

\subsection{\textbf{Motivation}}

Fig. \ref{fig:Motivation_Fig}a shows the inception (3b block) architecture from GoogLeNet\cite{GoogleNet}, a widely-used CNN model. Starting with Pool1 layer as input, it has four parallel computation paths, consisting of several 1x1, 3x3 and 5x5 CONV layers, output from which finally gets concatenated.
Fig. \ref{fig:Motivation_Fig}b represents the sparsity associated with different layers across both feature map and error gradients in GoogLeNet.
Note that when reporting sparsity of a layer, we consider the sparsity of output of a layer in either the forward or backward pass. For example, the $(f2,g1)$ pair in Fig.~\ref{fig:Motivation_Fig}c are the outputs of the CONV layer in forward and backward passes. The sparsity in $f2$ and $g1$ are reported as feature and gradient sparsity for the CONV layer, respectively.

We derive the following two key insights : First, \textbf{there exists significant sparsity in the feature maps and gradients which varies from $\approx$25\% to $\approx$55\%} in our observed benchmarks. Similar levels of sparsity have also been reported in prior works \cite{albericio2016cnvlutin, snapea, sen2018sparce}. In general, we observe that sparsity in CNNs ranges from ~30\% to ~70\%, which is shown in Fig. \ref{fig:Motivation_Fig}d, for the five CNNs. 
Second, sparsity is present only across the ReLU layers and most importantly, \textbf{the sparsity is identical for the feature maps and error gradients}. 
We analyze this unique relationship in section \ref{sec:why_sparsity}
and provide mathematical formulation behind the same.

Now, computing a single output neuron value (feature/gradient) can take several thousands of multiplications. For example, a layer with input dimension ($[C,H,W]$, C: input channel, H:input height, W :input width), filter dimension($[M,C,R,S]$, M: \#Filters, R: Filter Height, S: Filter Width), and output dimensions($[M,U,V]$, U:O/p height, V:O/p width) expressed as: $[C,H,W]$ $\xrightarrow[\text{}]{\text{[M,C,R,S]}}$ $[M,U,V]$ consumes $C\times R\times S$ number of Multiply and Accumulate (MAC) operations to produce a single output value. The total number of computations required is thus given by $M\times U\times V$ $\times$ $C\times R\times S$. 
In general, if output has a sparsity fraction of $s_{f1}$, total amount of computation necessary is proportionally reduced by factor of $(1-s_{f1})$. Also, the proposed approach is complementary to input sparsity, therefore, having a sparse input fraction of $s_{f2}$ can further reduce computation down to $(1-s_{f1})$ $\times$ $(1-s_{f2})$. \textit{Note that the levels of sparsity observed is true for every single iteration of the training step, for every single CONV-ReLU pair present in the network}. 
Since an ImageNet-scale network training can span millions of steps, this presents a unique opportunity for reducing the cost of training if the inherent sparsity is exploited.

In addition, there are also certain network structures that only allow for exploitation of output sparsity in the backward pass. This is due to the introduction of batch-normalization (BN) layers between CONV and ReLU pairs. An example is shown in Fig.~\ref{fig:Motivation_Fig}c. 
Here the input sparsity is applicable when considering \textbf{f1} as input to the CONV layer (\textbf{f1} is preceded by ReLU layer and is sparse). 
However, input to the same CONV layer \textbf{g2} in backward pass is not sparse, as the gradient \textbf{g3} gets re-normalized after passing through the BN layer. However, computation of \textbf{g1} can still allow output sparsity, as \textbf{g0} is sparse due to the ReLU layer. 
Note that BN is a popular approach in network training as it leads to faster convergence, and therefore, in such scenarios straightforward way of adopting input sparsity in the backward pass is not applicable. This further underlines the \emph{importance of output sparsity} and the need for efficiently leveraging them during gradient computation stage. 

\textcolor{blue}{It is also important to note that recently proposed \textit{Swish} \cite{ramachandran2017searching} non-linearity (also used in EfficientNet \cite{tan2019efficientnet}) tends to provide slightly better accuracy than ReLU (within 1\%). However, Swish  does not lead to a ``direct" sparsity of activation and gradient values, unlike ReLU. Therefore, ReLU can still be treated as an efficient approximation of Swish, involving trade-off between accuracy ($<$1\%) and speed-up (up to 2x). The proposed scope of this work is thus limited to using ReLU based non-linearity as activation function.}

\subsection{\textbf{Related Work}}
At a high level, sparsity exploitation techniques can be classified mainly as \textit{input} or \textit{output} type. A given technique is said to be input sparsity when it skips computation corresponding to zero values of either or both the input operands.
Examples of input sparsity during forward pass include Cambricon-X \cite{cambricon-x} which performs computation skipping on zero valued weight parameters. CNVLUTIN \cite{albericio2016cnvlutin} performs dynamic skipping of neuron activation by run-time encoding and skipping of zero valued input neurons. Recently proposed SCNN \cite{scnn} architecture jointly takes advantage of weight and activation sparsity during inference phase.

The Backward Pass (BP) of neural network training can be considered similar to Forward Pass (FP), with weight parameters transposed and activation map replaced with corresponding gradient values. Therefore, the above scope is equally applicable in the backward pass of training, subject to sparsity of involved operands.
Several recent works have been proposed in the context of training \cite{eagerpruning, spartann, sparten, sparsetrain} leveraging sparsity in different phases of training (FP, BP and WG).
However, all such works are limited by using only input sparsity during gradient back-propagation. This also has an important implication for the networks with batch normalization (BN) layer. With the BN layer included, gradients are \emph{no longer sparse} for input sparsity exploitation. Thus, all prior works either use networks which appeared before BN (Alex, VGG, Google as in \cite{spartann, sparten, rajbhandari2017optimizing}) or simply assume that a BN layer is not necessary in the model to help retain input gradient sparsity \cite{sparsetrain}. The novelty of our technique lies in its ability to leverage gradient output sparsity despite the presence of a BN layer, and otherwise, jointly leverage input and output sparsity.  \textcolor{blue}{Lee et al \cite{lee2019acceleration} proposes to skip computation of gradient values for non-BN networks based on ReLU activation, however it doesn't consider input sparsity of activation (during FP) or gradient values(in BP or WG). Another closely related work \cite{dac_paper} proposes to leverage gradient output sparsity, however, it lacks key algorithmic (i.e., why and what exactly makes gradients sparse) and architectural insights (i.e., issues in designing a sparse accelerator, load imbalance, retaining high PE utilization, data reuse) to leverage input and output sparsity efficiently. Liu et al. \cite{DSG} propose to estimate critical neurons in the output layer prior to their computation, thereby, constructing a sparse network execution graph during each iteration of training. Although this is akin to leveraging the notion of output sparsity starting from FP itself, the scope of ReLU-driven sparsity for gradients during BP still remains valid for the set of retained neurons and our proposed technique remains applicable.}
                                            
Note that the scope of weight sparsity during training is the result of pruning, which introduces opportunity for leveraging weights and activation (gradient) sparsity simultaneously \cite{eagerpruning}.  
While pruning is an algorithm level optimization and is orthogonal to our proposed approach, it can still be supported by providing additional hardware units. 
However, we limit the scope of this work to only non-pruned training, and focus on the dynamic sparsity, specifically, leveraging gradient output sparsity.

\section{Sparsity Characteristics}\label{sec:why_sparsity}

In this section, we discuss two fundamental attributes associated with neural network sparsity. First, we discuss why operands are sparse in a CNN network. Second, we analytically show that a ReLU layer forces identical sparsity footprints for the gradients in the backward pass as the output feature map sparsity in the corresponding (previous) forward pass.



\subsection{\textbf{Types and Sources of sparsity in CNNs}}

Sparsity in deep neural networks can be classified as static or dynamic. Static sparsity is associated with weight parameters that are obtained by performing quantization or thresholding as a post-training optimization step.
On the other hand, CNNs exhibit two types of dynamic sparsity at run-time: Sparsity of the feature map values during FP and sparsity of neuron gradients during BP. 
Two factors which contribute to dynamic sparsity during a neural network training step are weight distribution and input pre-processing. Typically, neural network weight parameters are initialized following a normal distribution around zero mean and the distribution remains so even for a fully trained network. In addition, neural networks require raw data to be pre-processed via input normalization, which
essentially results in a zero mean distribution of input feature values, indicating the potential for pre-activation values to be negative after performing weight multiplication and summation. Thus passing these pre-activation values through a ReLU layer produces zero output for the negative activation values. This effect ripples from layer to layer, continuing until the end of network. Similar phenomenon also occurs for the back-propagated gradient values. As we will see next, sparsity of feature values directly influences sparsity of gradient values passing backward through the ReLU layer and sets up the premise for our proposed optimization.

\subsection{\textbf{Why Sparsity is identical across ReLU Layer during a Forward and a Backward Pass?}}\label{identical_sparsity}

In order to understand the unique characteristics of the ReLU layer, we look at the execution of a sample training step consisting of forward and backward passes. 
For simplicity, we use a simple MLP neural network shown in Fig.~\ref{fig:MLP}, but the methodology used here can be extended to CNNs as well. 
Our example MLP consists of an input layer, a set of hidden layers and an output layer. 
Usually when describing neural networks, it is common to describe the weight and activation layers sandwiched together in the same layer. 
However, to understand the identical sparsity pattern in FP and BP, we will consider weight and activation layers separately. 
We denote the activation of a neuron $'j'$ in layer $'l+1'$ by $a^{l+1}_j$, which is obtained by applying a transfer function $\sigma$ to the accumulated sum  $z^{l}_j$ from the previous layer. $a^{l+1}_j$ is expressed as:

\begin{figure}[h]
    \centering
    \includegraphics[scale=0.22]{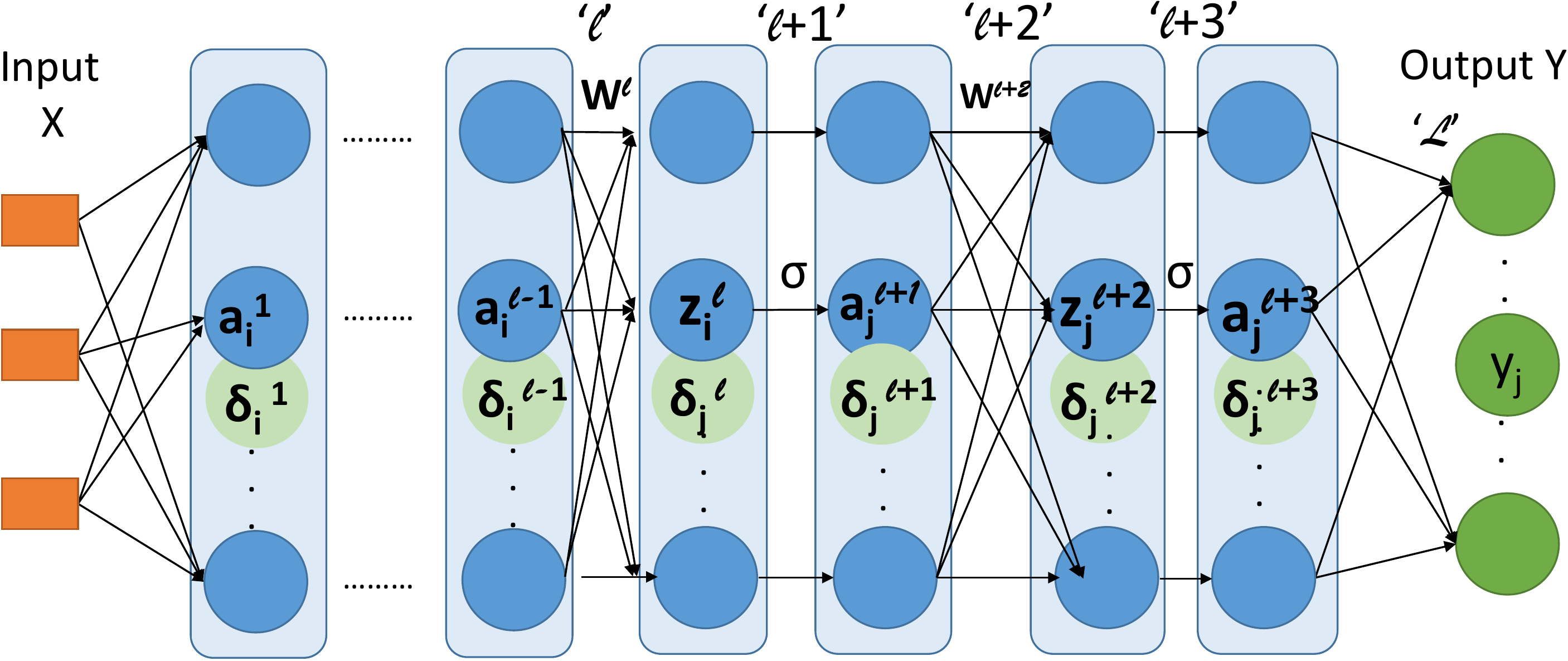}
    \caption{A simple MLP network to illustrate the notion of backpropagation.}
    \label{fig:MLP}
\end{figure}

\begin{figure*}[t]
    \centering
    \includegraphics[width=0.8\linewidth]{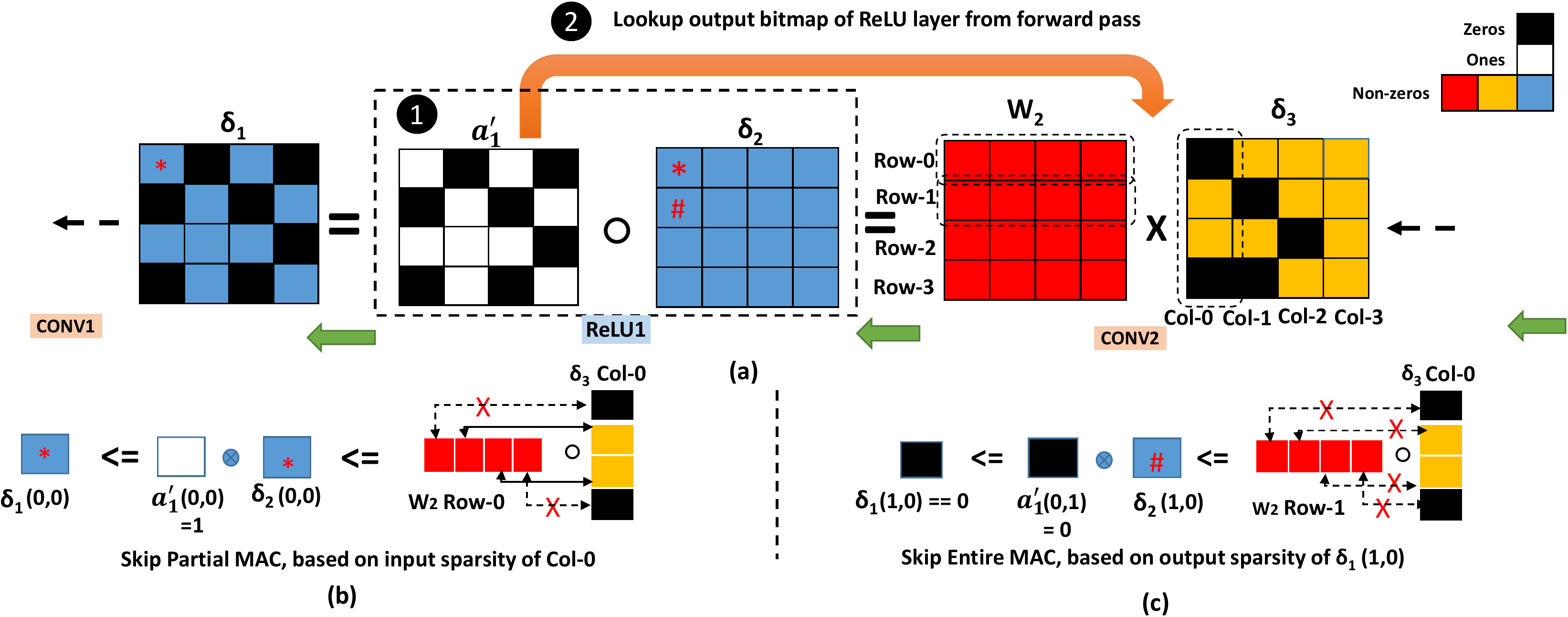}
    \caption{(a) Error gradient passing through Conv1-Relu1-Conv2 chain in backward pass (b) Multiplying $W_2$ Row-0 and $\delta_3$ Col-0 requires 2 MAC operations, considering input sparsity of $\delta_3$ Col-0 (c) Multiplication involving $W_2$ Row-1 and $\delta_3$ Col-0 can be entirely skipped considering output sparsity of $\delta_1$ matrix at index (1,0) location (value of $a_{1}^{\prime}$(1,0) is known beforehand).}
    \label{fig:Compaction}
\end{figure*}

\begin{align*}
    a^{(l+1)}_j = {\sigma(z^{(l)}_j)} , where ~z^{(l)}_j = \Sigma(w^{(l)}_{ji} a^{(l-1)}_{(i)}),
\end{align*}

    where, $w^{l}_{ji}$ is the weight value connecting the output neuron $j$ of layer $l$ to the input neuron $i$ of layer $l-1$.
    
The above equation is applied recursively from layer $l=1$ (first hidden layer) till the last layer $l=L$. For a labeled dataset $(X,Y)$, we compute the error at the last layer as a function of label output $y_j$, and activation at the last layer, $a^L_j$, which is denoted by a cost function $C$, where $C=f(y_j, a^L_j)$

The error gradient for the last layer is computed as follows: 
\begin{align*}
\delta^{(L)}_j = \frac{\partial C}{\partial {a^{L}_j}} = f^{\prime}(y_j, a^L_j).
\end{align*}

The error gradient for any layer $l$ (except the last layer) is computed by applying the chain-rule as below:  
\vspace{-2mm}    
    \begin{align*}
    \delta^{(l+3)}_j = \frac{\partial C}{\partial {a^{l+3}_j}} = \frac{\partial C}{\partial {a^{L}_i}} \times \frac{\partial {a^{L}_i}}{\partial {a^{l+3}_j} } =   \big[ \Sigma \delta^{(L)}_i w^{(L)}_{ij} \big].
    \end{align*}

Stepping back one more layer,
\vspace{-2mm}
\begin{align*}
\delta^{(l+2)}_j = \frac{\partial C}{\partial {z^{l+2}_j}} = \frac{\partial C}{\partial {a^{l+3}_j}} \times \frac{\partial a^{l+3}_j}{\partial {z^{l+2}_j}} = \delta^{(l+3)}_j \odot \large{\sigma^{\prime}(z_j^{l+2})}.
\end{align*}

As seen above, the error gradient at the output of transfer layer is given by the Hadamard product (element-wise multiplication) of error gradient of the subsequent layer and derivative of its own transfer function. In practice, when ReLU ($max(0, z_j^l)$) is used as the transfer function, the corresponding derivative is given by 



\quad \quad \quad \quad \quad
\boxed{
     \sigma^{\prime}(z_j^{l}) =
          \begin{cases}
           1       & \quad \text{if } z_j^{l} \geq 0\\
           0       & \quad \text{otherwise}.
        \end{cases}
    }
\vspace{1mm}

This means ReLU layer either retains (multiply with 1) or zeros out (multiply with 0) the gradient from the subsequent layer. 
Thus, the neuron location which will be zeroed out can be known apriori, since it is dependent on the value of the same neuron during the forward pass. 
Therefore, we can save significant computation by \emph{not computing gradients at those specific locations which will be zeroed out by ReLU in the backward pass}.


Fig.~\ref{fig:Compaction}(a) describes this situation: We represent the computations involving error gradient {\large$\delta_3$} and weight $W_2$ as a simple matrix-matrix multiplication. Following along the direction of green arrows, input matrices {\large$\delta_3$} and $W_2$ are multiplied together to generate the output matrix {\large$\delta_2$}, which is the output error gradient of the $CONV2$ layer. 
Next, a Hadamard (element-wise, point-to-point multiplication) operation is performed between elements of {\large$\delta_2$} and  ${\large{a_1^{\prime}}}$ to obtain {\large$\delta_1$}({\large$\delta_1$} $\xleftarrow{a_1^{\prime}}$ {\large$\delta_2$}). 
Given that incoming gradient values corresponding to {\large$\delta_3$} can themselves be sparse (0s shown as black squares in Fig.~\ref{fig:Compaction}), we can exploit the notion of input sparsity here by eliminating zero valued input operands of {\large$\delta_3$}. Therefore, multiplying Row-0 of $W_2$ with Column-0 of {\large$\delta_3$}, we only perform 2 MAC operations, instead of 4, to generate {\large$\delta_2$}(0,0), which is multiplied with ${\large{a_1^{\prime}}}$(0,0) = 1 to finally generate {\large$\delta_1$}(0,0) (= {\large$\delta_2$}(0,0)).

\begin{figure*}[t]
\centering
\includegraphics[width=0.9\linewidth]{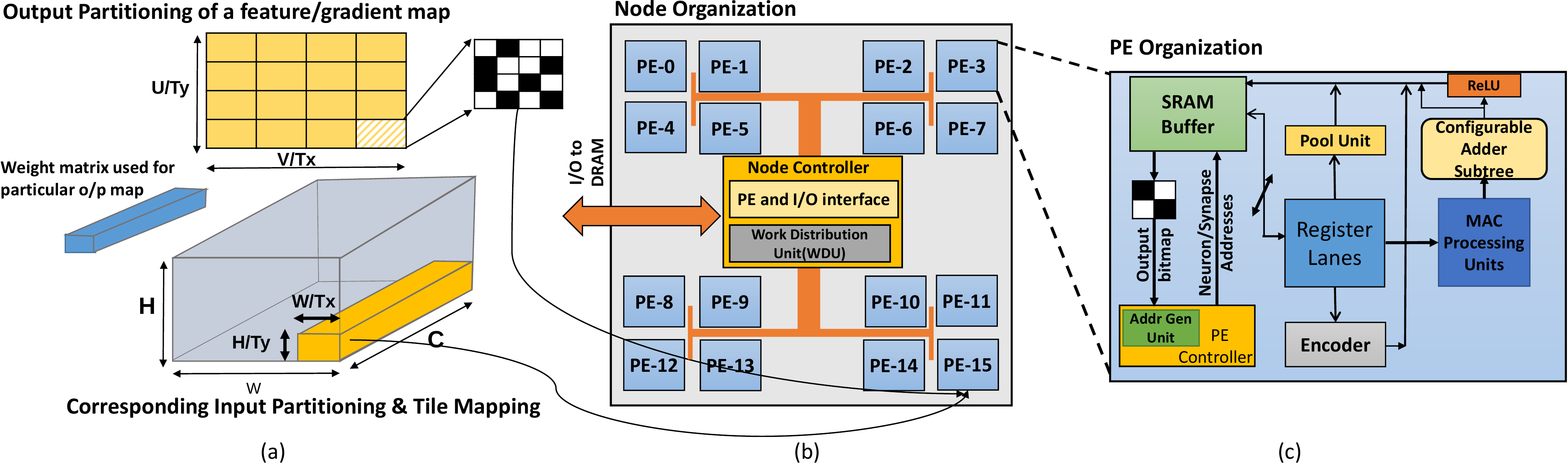}
\caption{Overview of the proposed architecture: (a) Input placement and mapping strategy; (b) Node Architecture consisting of a 2D grid of PEs for $T_x=4$ and $T_y=4$(Note that design can be scaled by increasing $T_x$ and $T_y$ values) (c) PE Schematic showing local SRAM buffer, PE-controller, Reg-array, MAC, Pool, Encoder and adder tree}.
\label{fig:proposed_design}
\vspace{-4mm}
\end{figure*}
Notice however that, unlike  {\large$\delta_1$}(0,0), the entry at location {\large$\delta_2$}(1,0) eventually gets zeroed due to multiplication with ${\large{a_1^{\prime}}}$(1,0).
Therefore, we can entirely avoid computing {\large$\delta_2$}(1,0) based on the notion of output sparsity, as shown in Fig.~\ref{fig:Compaction}(c), thus leading to further savings in terms of MAC operations. 
Thus, exploiting output sparsity offers significant opportunity towards computation reduction, hence gain in performance and/or energy efficiency of execution. In addition, the flexibility of this approach leaves ample room for us to jointly exploit input and output sparsity simultaneously thus reducing the number of ineffective computations even further. 

\textbf{Why an accelerator-centric approach?} 
CPU/GPU based SW implementations are capable of exploiting input sparsity only at very high sparsity levels (90\%)\cite{sen2018sparce}. This is also reported in recent accelerator based sparsity works \cite{sparten, spartann, albericio2016cnvlutin}. In addition, our optimization requires output sparsity exploitation which is not supported by existing library frameworks. Our own CPU and GPU implementations indicate that required break-even sparsity levels are (~90\%), significantly higher than actually observed in the training context. 


\section{\textbf{Design of the Proposed Architecture}} \label{section: design}
In this section, we describe the design details of our proposed hardware architecture. 

\subsection{\textbf{Overall Organization \& Operation}} 


Fig.~\ref{fig:proposed_design} shows a schematic of the proposed micro-architecture. The basic building block of the unit is a Processing Element (PE). 
A PE consists of specialized register array organized in a number of lanes, each with a 16 bit floating point (FP) multiply \& accumulate unit and a re-configurable adder tree shown in Fig.~\ref{fig:proposed_design}c. 
The register array also feeds a pool and encoder unit (discussed later). The PE also contains a local SRAM buffer for storing weights and neuron values pre-fetched from memory and an input address generation unit. As shown in Fig.~\ref{fig:proposed_design}b, a single node has a number of such PEs, which interact with a Work Distribution Unit (WDU) via an H-tree-based interconnect, located within a central node controller unit.


\subsection{\textbf{Computation Placement and Mapping}}
Consider forward pass of a CONV layer given by: $[C,H,W]$ $\xrightarrow[\text{}]{\text{[M,C,R,S]}}$ $[M,U,V]$ (for notations, refer to Section~\ref{sec:bgmv}). The backward pass of the same layer can be given as $[C,H,W]$ $\xleftarrow[\text{}]{\text{[C,M,R,S]}}$ $[M,U,V]$. Note that, FP and BP share the same computational property of a CONV operation, with $M$ and $C$ parameter interchanged.
In the following discussion, we will use the forward pass notation, but output sparsity is applicable only during the backward pass. Also, we will be using the terms filter, kernel or synapse interchangeably to refer to the weight parameters of a layer.

Each of the PEs in our design is assigned to compute a fragment of the output feature map given by the size ($\frac{U}{T_x} \times \frac{V}{T_y}$). To support this computation assignment, the input feature map 
is tiled into $T_x \times T_y$ fragments, where $T_x$ and $T_y$ refer to the number of PEs along the horizontal and vertical dimensions, respectively. This is shown in Fig.~\ref{fig:proposed_design} (a) and (b). Thus, the size of each input tile fragment is given by $C \times (\frac{H}{T_x} + 2 \times \lfloor R/2 \rfloor)  \times (\frac{W}{T_y}+ 2 \times \lfloor S/2 \rfloor)$. The extra $(R/2, S/2)$ term is factored in towards \textit{input Halo} resolution~\cite{scnn}. Similarly, at the layer boundaries of execution, the extra amount of data is exchanged between the neighboring tiles towards resolving \textit{Output Halo} condition. A single weight kernel is streamed in from DRAM into the Node, and broadcast to all the PEs. Once a filter has finished executing on the input fragment, the next filter is loaded into the PEs and the computation proceeds, until all the $M$ filters of the current layer have finished execution. 
For WG computation stage, we follow similar partitioning strategy with respect to activation and neuron gradient maps. However, the data-access pattern is different as compared FP and BP stages, which is performed accordingly.

\begin{figure}[h]              
\centering                                   
\includegraphics[width=0.9\linewidth]{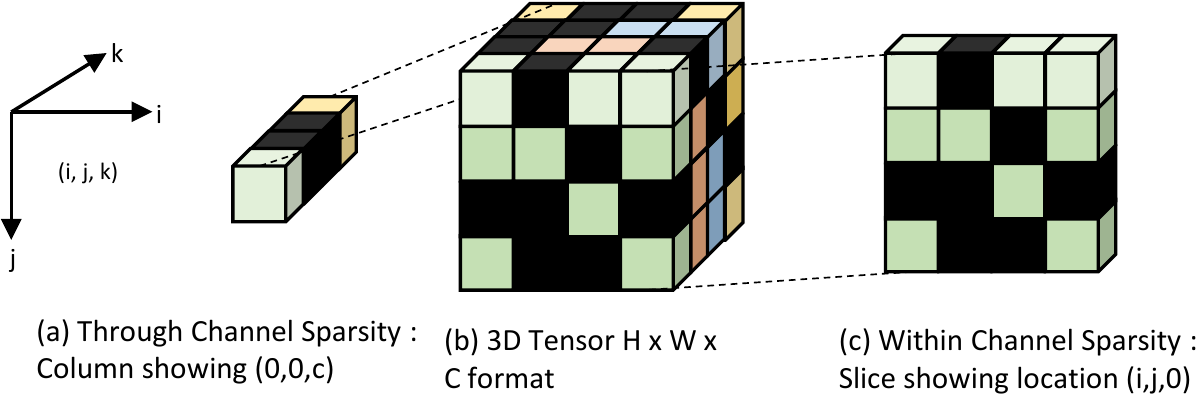}
\caption{Viewing sparsity from a multi-dimensional tensor perspective (zero values are shown in black color).}
\label{fig:ACTC-Sparsity}
\end{figure}

Note that the proposed architecture takes advantage of both input and output sparsities of feature maps and error gradients. Before delving into the details of the architecture, we revisit the concepts related to sparsity, which are essential to our design, with reference to a 3-dimensional feature map representation (given by $C \times H \times W$) discussed next. 

\textbf{Through Channel (TC) Sparsity:}
This is the sparsity that is associated with every neuron location ($H_i, W_j$) in a 2D feature map along the channel dimension. When working with a 3D tensor representation of feature maps, $R^{C \times H \times W}$, through channel sparsity can be defined as the set: 
$$ S_{H_iW_j \in {H \times W}} = \{ x | x = 0 , x \in \left[ C_k, H_i, W_j \right], \forall C_k \in {C} \}$$ 
\vspace{-2mm}

The notion of TC sparsity is important while we seek to leverage the input sparsity of a $CONV$ layer, where the filter map ($C \times R \times S$) performs element-wise multiplication through the channel depth ($C$) of a $R \times S$ sized two dimensional area.

Fig.~\ref{fig:ACTC-Sparsity}a shows the concept of through channel sparsity. One of the key requirements to translate sparsity into performance improvement is to not spend computational cycles in identifying zero values during execution. Therefore, this necessitates an indexing stage that can identify non-zero (NZ) neuron locations.
\begin{figure*}[t]
   \centering
	\begin{subfigure}{0.35\textwidth}
	   \centering
	   \includegraphics[width=1\textwidth]{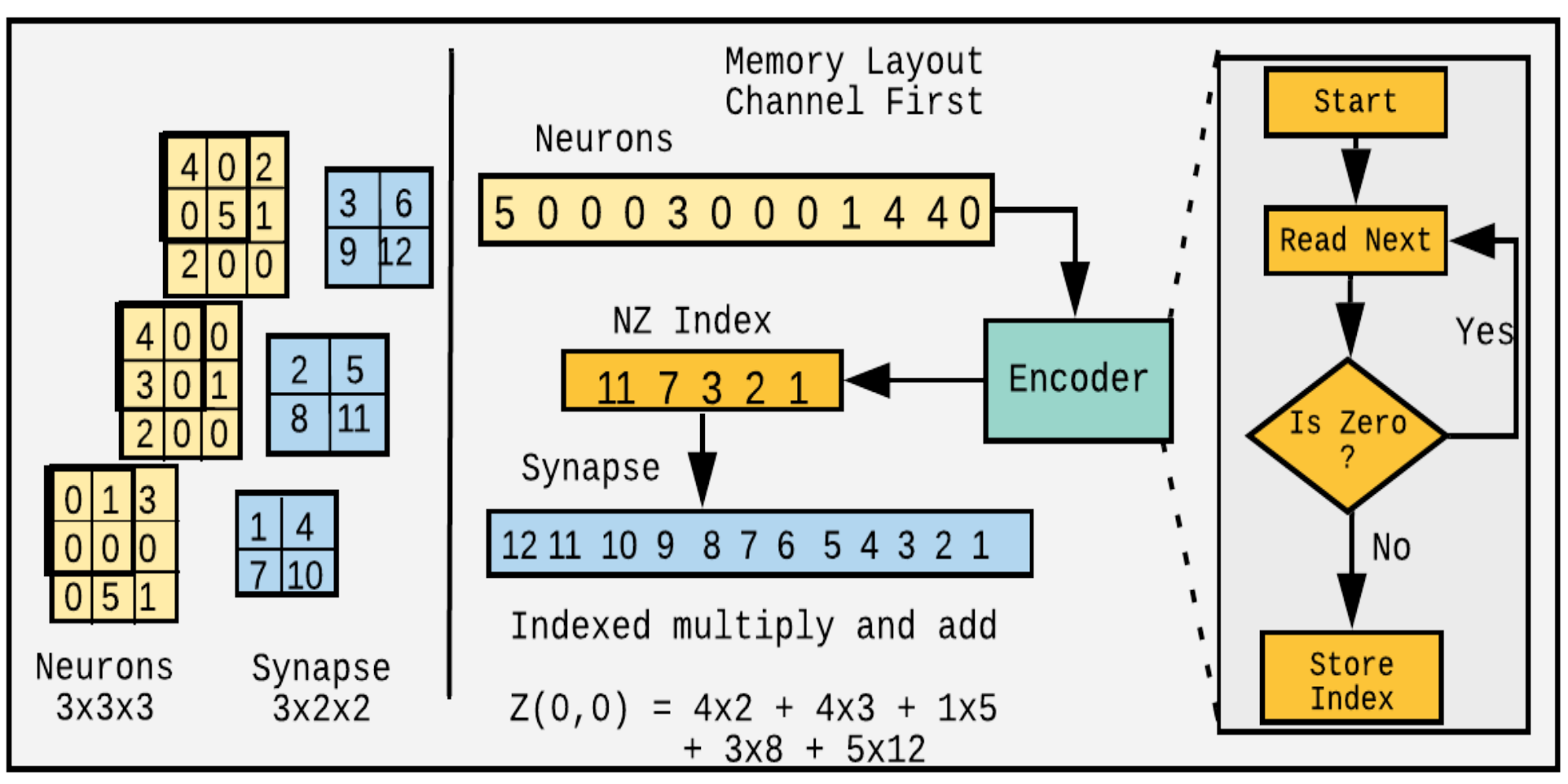}
        \caption{Input Sparsity with neuron indexing}
        \label{fig:Input_Sparsity}
    	\end{subfigure}
    \quad
  	\begin{subfigure}{0.25\textwidth}
	   \centering
	   \includegraphics[width=\textwidth]{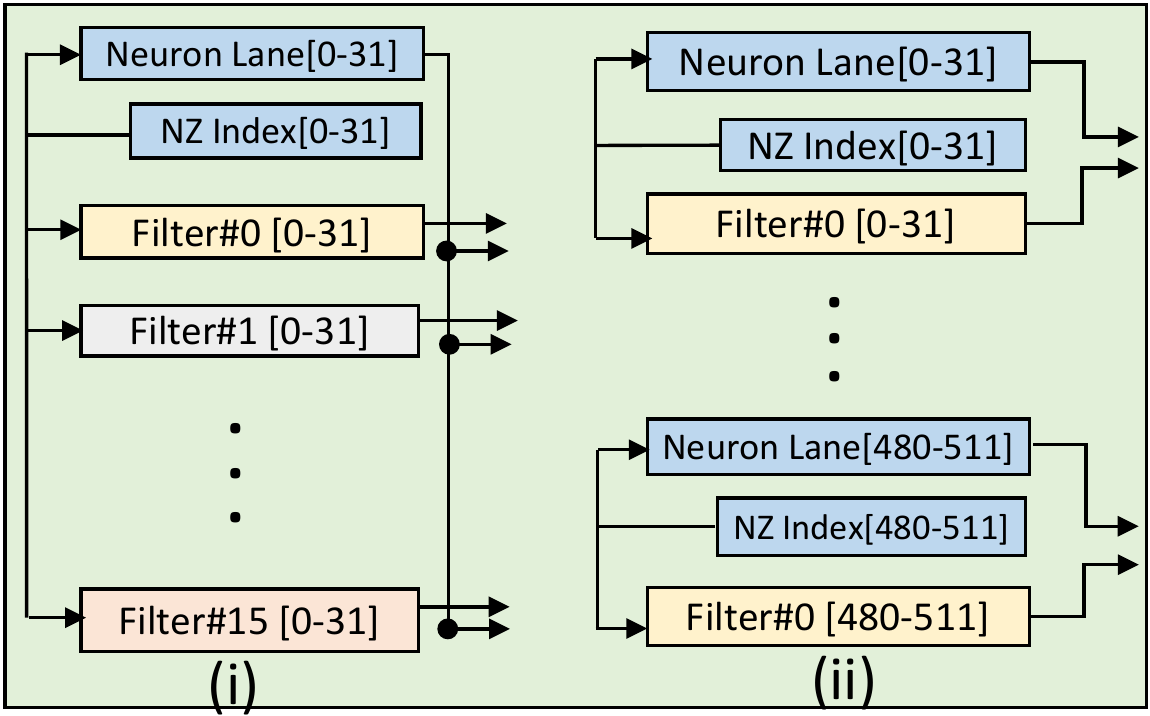}
	   \caption{Filter Decoupling}
	   \label{fig:filter_decouple}
	\end{subfigure}
    \quad
	\begin{subfigure}{0.33\textwidth}
	   \centering
	   \includegraphics[width=\textwidth]{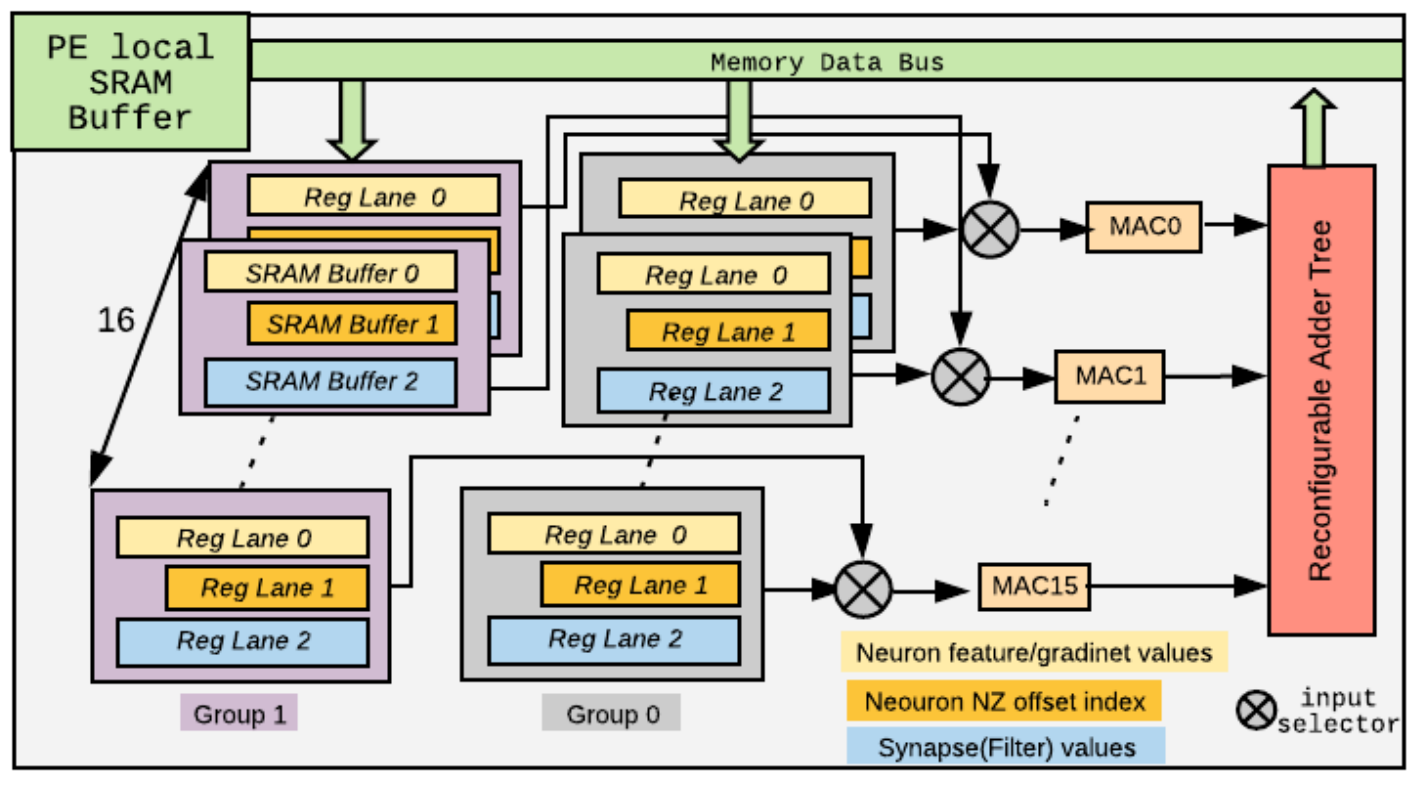}
	   \caption{Computation lanes with double buffering}
	   \label{fig:PE_micoarch}
	\end{subfigure}
	\vspace{-3mm}
\caption{Architectural design considerations}
\end{figure*}

In our design, we perform NZ indexing of the generated feature/gradient map at the end of completing all the kernel execution of a given layer. This indexing is performed through channel dimension of the output feature/gradient map, for a length of 32 at a time, and its output represents offset indices corresponding to the non-zero neuron locations, stored as an offset map.
Notice that indexing only needs to be performed once per layer, and the indexed neuron values are heavily reused subsequently ($ O(M \times k^2) $) which is 2-3 orders of magnitude, the latency and energy cost of encoding is well amortized. We also retain the memory access regularity by only indexing the neurons (not compression).



Fig. \ref{fig:Input_Sparsity} describes this idea for an example feature map of size 3x3x3 and a synapse size of 3x2x2, by showing the computation flow for the first output location denoted as $Z(0,0)$. 
Note the memory ordering of the synapse values and corresponding neuron values, which is stored according to "channel first" layout. 
For the neuron values, we also extract the NZ index locations using the Encoder.  
The NZ index values are used to index into the synapse field, and thereby selective (only non-zero) neuron and synapse multiplication takes place and final sum is accumulated (Z(0,0)). 

\textbf{Within Channel (WC) Sparsity:}  
Within channel sparsity is defined as the sparsity associated with each channel of the feature maps, obtained by looking at all the $H \times W$ 2D neuron values that are zeros, contained in a particular channel $C_k$.  Mathematically, this can be defined as the set,\\\null\vspace{-6pt}\\ 
 \vspace{2pt}
 $S_{C_k \in {C}} = \{ x|x = 0, x \in [C_k, H_i, W_j], \forall (H_i, W_j) \in {H \times W}\}$
 \vspace{4pt}
 
\begin{figure}[h]
\centering
\includegraphics[width=0.45\textwidth]{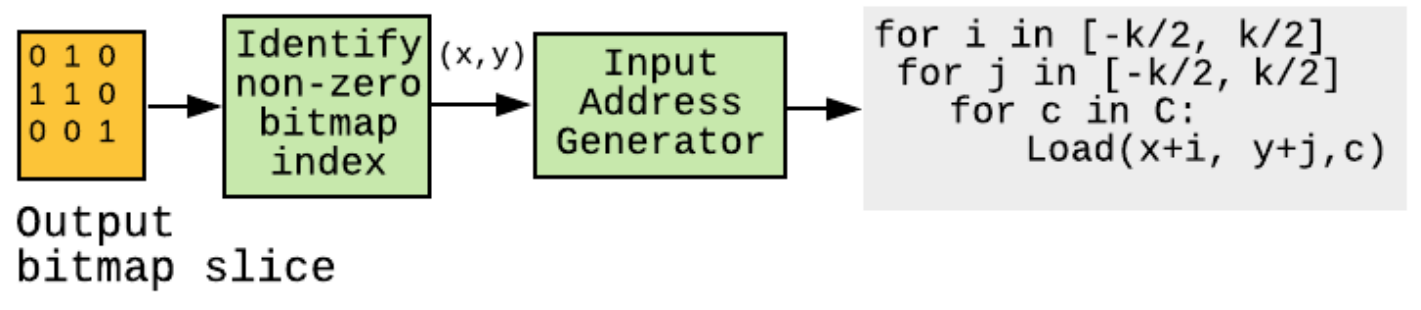}
\caption{Supporting Output Sparsity: Address generator module loads input only according to the non-zero output bitmap co-ordinate.}
\label{fig:Output_Sparsity}
\end{figure}

Fig.~\ref{fig:ACTC-Sparsity}c shows a representative example of WC sparsity. The notion of WC sparsity is important while exploiting the concept of output sparsity . The sparsity within each channel determines which output neurons can be skipped during the backward pass of a training step. Therefore, as shown in Fig~\ref{fig:Output_Sparsity}, input address generator skips all the zero valued bitmap locations and loads input only corresponding to non-zero bitmap co-ordinate. Furthermore, if the input was also sparse, it will also load the offset index values and computation can proceed, taking advantage of TC sparsity.


\textbf{Filter Decoupling:}
This leads us to another important aspect of accelerator design choice, which is filter decoupling. In order to maximize the parallelism of the execution, designs tend to couple multiple filter lanes with a shared neuron lane. 
This is shown in Fig 8.(b) (i), where a single neuron lane and it's NZ offset index is broadcast to multiple filter lanes and can effectively exploit input sparsity.
However, for exploiting output sparsity, the same approach cannot be adapted.
This is because different channels of output gradient map would have different sparsity footprints, the corresponding filters need to process different output index locations at any given point in time. Computing different output index locations requires different region of input neurons to be present on the computation lanes. Hence, it is necessary to decouple filter lanes corresponding to individual filters/gradient maps(either spatially or temporally). Accordingly, all available lanes are dedicated to produce o/p corresponding to same filter or gradient map at a time, allowing for highly streamlined SRAM memory access pattern and sustaining a high bandwidth. This is shown in Fig.8b(ii). 
This decoupling is essential in exploiting output sparsity jointly with input sparsity and is a distinguishing architectural feature of our design.






\vspace{-1mm}
\subsection{\textbf{Node Micro-architecture}} 

Each PE in our node design consists of 16 computational lanes, each consisting of 3 separate SRAM buffers and a Multiply and Accumulate (MAC) unit. 
The three buffer arrays correspond to the neuron lane, the offset lane and the synapse lane, shown in Fig.~\ref{fig:PE_micoarch}, which are again divided into two groups (group 0 and group 1) for the purposes of double buffering. 
Within a group, each buffer lane can hold 32 entries, where each entry in the neuron and synapse lanes is 2 Bytes in size and each offset lane entry is 5 bits to hold the non-zero index values corresponding to 32 entries. Thus, size of neuron and synapse lanes are 64B, and offset lane has a size of 20B(5 bits x 32). At each cycle, the on-chip SRAM memory needs to deliver 64B of neuron and 20B of offset values, thus required maximum bandwidth is given by 84Bytes/cycle. 
After loading data onto the lanes, offset lane is read sequentially and output of the offset lane indexes into the neuron and synapse lane, to obtain the non-zero neuron and corresponding weight value. 
The associated MAC unit receives the neuron and synapse values and updates its accumulated sum after performing the multiplication. 
The outputs from all the MAC units are connected to a re-configurable adder tree where it can reduce its operands from the 16 data lanes to a single value. 
The reduction takes place once all the non-zero neuron values have been processed in all the computation lanes in the current group. 
Due to a possible un-even distribution of non-zero values in each lane, there can be lane stall cycles where computation in a lane cannot proceed and needs to wait for other lanes to finish. This issue is addressed by double buffering inputs to the computation lanes, through which probability of lane stall is effectively reduced. 

An important point to note here is that each PE can hold a maximum of 1024 pairs of input entries (16 lanes x 32 entries/group x 2 group) corresponding to the output neuron value computation. This configuration is ideal if the receptive field size (denoted as $ CRS = C \times R \times S$) of a given layer's filter is exactly 1024. However, all real world benchmarks have variable sizes of receptive fields within different layers that are not equal to 1024, and therefore, it has the following two consequences: first, memory bandwidth utilization and second, in terms of overall PE lane utilization. We discuss both the issues and corresponding optimizations that allow PEs to perform computation in a more efficient manner.
\subsection{\textbf{Synapse Blocking for Improved Memory Bandwidth Utilization ($CRS > 1024$)}} \label{section: blocking}
When the receptive field size of a filter is greater than 1024, a straightforward approach to follow is by finishing the entire required computation for each output in a single pass. However, this leads to inefficiency as synapse (filter) values have to be loaded and reloaded for every subsequent neuron output computation. Here, we address this issue by blocking the synapse values at 1024 element boundaries and generating partial sum for each output neuron location. In the next iteration, the remainder of the synapse values are blocked and the partial sum from previous iteration gets reduced with the computation currently being performed. The number of iterations required per PE is thus given by $CRS/1024$. 

\subsection{\textbf{Hierarchical reconfiguration with Re-configurable Adder Tree Structure ($CRS < 1024$)}} \label{section: adder_tree}

\setlength{\columnsep}{10pt}%
\begin{wrapfigure}{r}{0.25\textwidth}
    \centering
    \includegraphics[width=.25\textwidth]{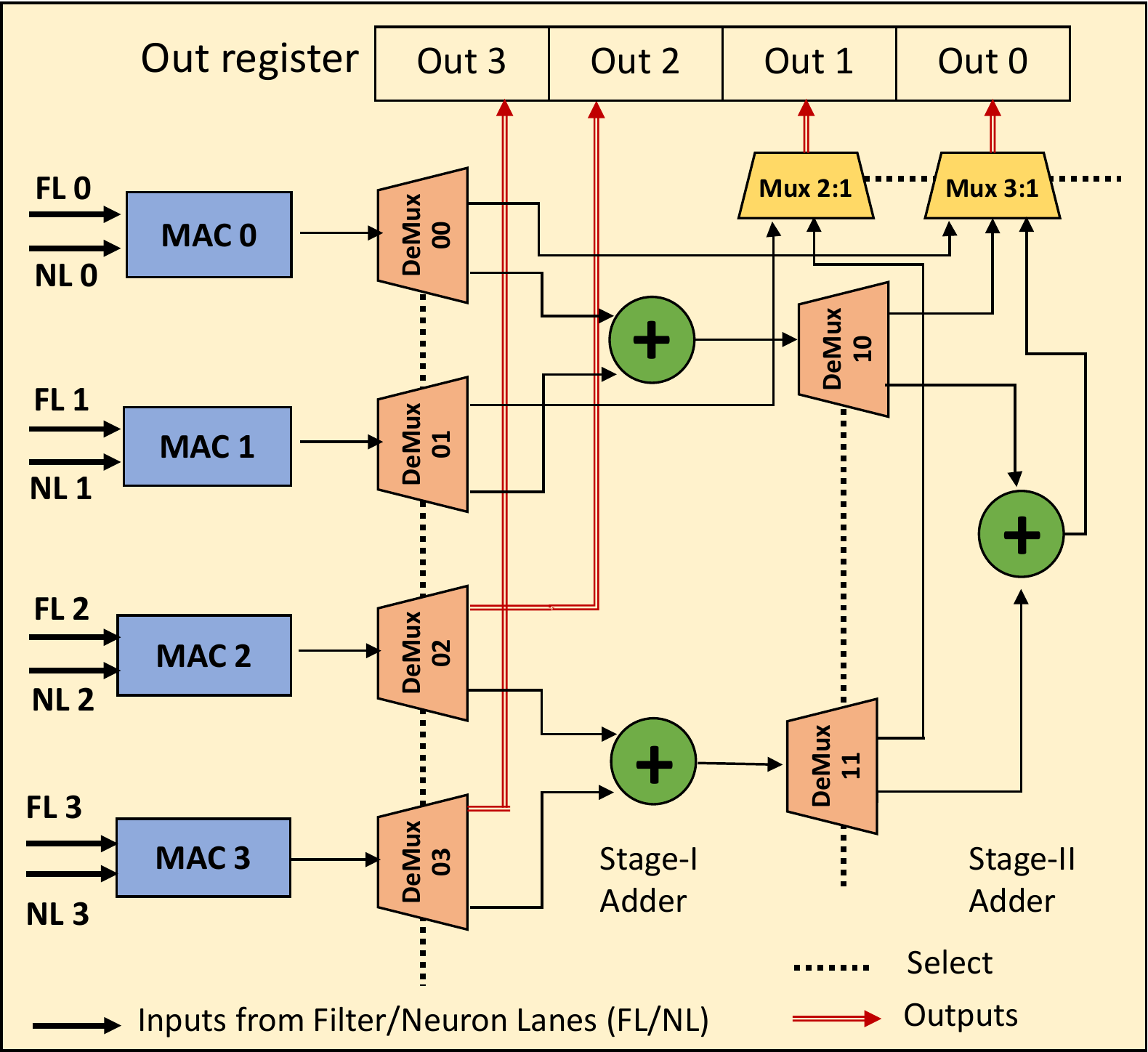}
    \caption{Re-configurable adder tree to support smaller receptive field sizes ($<$ 1024).}
    \label{fig:rec_field}
   
\end{wrapfigure}

In the case where receptive field size is less than 1024, it can lead to PE under-utilization
e.g. in the worst case scenario when $CRS <= 32$, the PE lane utilization drops to 1/16 or 6.25\%.

To achieve an efficient lane utilization, we propose a re-configurable adder tree that can selectively reduce output from the MACs (\textbf{in powers of 2}), and thus, enables multiple output neuron computations to proceed simultaneously, as shown in Fig. \ref{fig:rec_field} (for a lane count of 4). Similar to data flow control using multiplexers, we insert  de-multiplexers between successive adder stages. The demuxes are pre-configured to either forward input data to the next adder stage or re-route for memory storage. Accordingly, multiple independent output computation can proceed in parallel based on the size of the receptive field, leading to higher PE utilization.

To further alleviate the issue of under-utilization when lane occupancy is non-aligned (e.g., with occupancies such as 3, 5, 9, etc.) we propose to block the filter kernels to it's nearest aligned size (smaller than the required occupancy), and scheduling the remaining computations in the next iteration. In each iteration, lane sizes are recursively aligned. Thus, using hierarchical reconfiguration approach, we can achieve full utilization of PE, ensuring high performance.

\vspace{-0.5mm}
\subsection{\textbf{A Work Redistribution Strategy}}
\vspace{-0.5mm}
As discussed earlier, each PE is assigned to compute a region slice of $U/T_x \times V/T_y$ of the output tensor. 
During the backward pass, each PE is also provided with the corresponding output bitmap slice already stored in DRAM from the forward pass.  
However, it is observed that there exist variations within spatial sparsity distribution of a map, leading to some of the PEs finishing early and remaining idle. 
This limits the achievable speedup by the tile region having maximum amount of work. 
In order to avoid such a scenario, we propose to re-distribute work from a tile with the maximum amount of work (maximum number of non-zero neurons), to a tile which has just finished executing all neuron locations. 
This maximizes the utilization of any idle/available PEs, and helps to achieve higher throughput. 

Work re-distribution is supported by a centralized work re-distribution unit within a node, which tracks the progress of any given PE-tile by a state tuple $<iter,x,y>$, where $iter$ represents the iteration value w.r.t. partial sum, and (x,y) represents the co-ordinates of the output neuron currently being processed. Each PE contains its own boundary condition for these parameters which act as the start and end markers. WDU detects an idle tile("source") when all the parameters have reached their end markers. To redistribute work, it selects a tile("target") which has the lexicographically smallest value of state tuple signifying that this tile has the maximum amount of remaining work.

Our approach divides the remaining work in two halves and re-assigns the lower half to the target tile. During this process, WDU sends commands to source tile to send the pertinent input portion to target tile, and also updates the start and end markers of both the tiles accordingly. This redistribution comes at an additional overhead of sharing input data between the PEs and merging output results, therefore, it is useful to work-redistribute only when remaining work is above a certain threshold. In our evaluation, we empirically identify a re-distribution threshold of 30\% as a lower bound, although it is possible to dynamically adjust this value for more optimal results.

\section{Evaluation Methodology}

\subsection{\textbf{Neuron Activation and Gradient Traces:}} To obtain the activation and gradient values of a neural network training iteration, we use the publicly available TensorFlow framework. We model 5 state-of-the-art CNNs in TensorFlow, namely, VGGNet, ResNet18, GoogLeNet, DenseNet and MobileNet, for training on the Imagenet dataset consisting over a million training images. Note that, to evaluate the performance of our proposed architecture, we need to obtain the activation and gradient traces in a layer-wise fashion. 
Therefore, we use Tensroflow training framework to obtain layer-wise gradient and activation values of the neurons which we use as input traces for our accelerator simulation framework.

\begin{table}[ht]
\begin{tabular}{|l|l|l|l|l|}
\hline
\multicolumn{5}{|c|}{Processing Element (PE)}
\\ \hline

\textbf{Component} & \textbf{Param} & \textbf{Spec} & \begin{tabular}[c]{@{}l@{}} \textbf{Power /} \\ \textbf{Energy} \end{tabular} & \begin{tabular}[c]{@{}l@{}} \textbf{Area} \\ \textbf{($mm^2$)} \end{tabular}                             
\\ \hline

\begin{tabular}[c]{@{}l@{}} Neuron/syn. \\ reg. file \end{tabular} & 
\begin{tabular}[c]{@{}l@{}} count\\ size \end{tabular} & 
\begin{tabular}[c]{@{}l@{}} 64 \\ 4 KB \end{tabular} & 20.1 mW & 0.3820 
\\ \hline

\begin{tabular}[c]{@{}l@{}} Non-zero \\ idx reg. file \end{tabular} & 
\begin{tabular}[c]{@{}l@{}} count\\ size \end{tabular} & 
\begin{tabular}[c]{@{}l@{}} 32 \\ 0.625 KB \end{tabular} & 3.44 mW & 0.0602
\\ \hline

\begin{tabular}[c]{@{}l@{}} Mul-acc \\ (MAC) unit \end{tabular} & \begin{tabular}[c]{@{}l@{}} count\\ size \end{tabular} & \begin{tabular}[c]{@{}l@{}} 16 \\ 16b FP \end{tabular} & 10.56 mW & 0.1235
\\ \hline

\begin{tabular}[c]{@{}l@{}} Reconfig. \\ adder tree \end{tabular} & 
\begin{tabular}[c]{@{}l@{}} count\\ size \end{tabular} & 
\begin{tabular}[c]{@{}l@{}} 15 \\ 16-input \end{tabular} & 5.5127 mW & 0.0803
\\ \hline

\begin{tabular}[c]{@{}l@{}} Non-zero \\ encoder \end{tabular} & count & 1 & 0.7714 mW & 0.0113
\\ \hline

Control & - & - & 2.0955 mW & 0.0313
\\ \hline

SRAM buff. & 
\begin{tabular}[c]{@{}l@{}} bank size \\ bank count \\ access time \\ line size \end{tabular} & 
\begin{tabular}[c]{@{}l@{}} 32 KB \\ 4 \\ 0.80787 ns \\ 128B \end{tabular} & 
\begin{tabular}[c]{@{}l@{}} 0.035 nJ/rd \\ 0.040 nJ/wr \\ 25 mW (D) \\ 8.1 mW (S) \end{tabular} & 0.3696

\\ \hline

\textbf{PE total} & - & - & \textbf{75 mW} & \textbf{1.0468}
\\ \hline \hline

\multicolumn{5}{|c|}{Proposed design node at 667 MHz at 32nm}
\\ \hline
PE & count & 16x16 = 256 & 19.2 W & 266.24
\\ \hline




\end{tabular}

   \caption{\textbf{Component specifications for our design.}}
   \label{tab:config}
\end{table}

\subsection{\textbf{Proposed Design Configuration and Simulation Framework:}} 
The individual components of our proposed architecture is modeled at the Register-Transfer level using Verilog and synthesized using the Synopsys Design Compiler with the Synopsys AED 32nm library. 
The components include a) half-precision MAC units, b) the re-configurable adder tree, c) the operand feeding register files, d) the non-zero activation encoding unit, e) the PE controller and f) other miscellaneous logic. The design specifications are reported in Table \ref{tab:config}. \textcolor{blue}{Note that we are using 16-bit floating point computation, which has been recently shown to achieve convergence on complex learning tasks using techniques such as loss scaling \cite{ginsburg2017training}. }  
We use CACTI \cite{cacti} tool to model and estimate the access parameters of on-chip SRAM buffers that feed neuron activation and synapse weight values to the 16 register lanes (Table \ref{tab:config}).
The area, power and timing estimates of all on-chip components obtained were then plugged in to our in-house cycle-accurate architectural simulator to obtain the energy and latency estimates of the different CNN workloads.
The simulation infrastructure models the events in terms of mapping and placement of different kernels involved in a typical neural network training process (i.e. activation forward pass, weight update and gradient backward pass) as various compute, memory and other on-chip transactions.
The 256 unit PE clusters with MAC and re-configurable adder tree units, provide a peak throughput of 8192 half-precision FLOPs/cycle (5464 GFLOPs/s).
The PEs communicate with each other through an H-tree on-chip interconnect equipped with a broadcast bandwidth of 512 GB/s.

\begin{figure}
    \centering
    \begin{subfigure}{0.45\textwidth}
      \centering
        \includegraphics[width=0.8\linewidth]{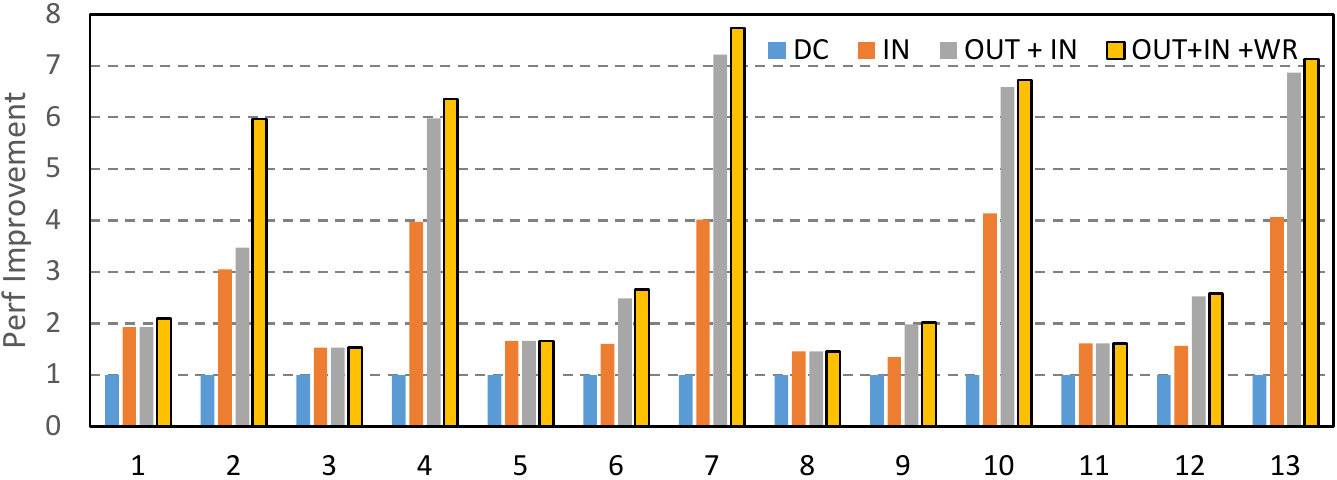}
        \caption{VGGNet}
        \label{fig:VGGNetLayerwise}
    \end{subfigure}%
    
    \begin{subfigure}{0.45\textwidth}
      \centering
        \includegraphics[width=0.8\linewidth]{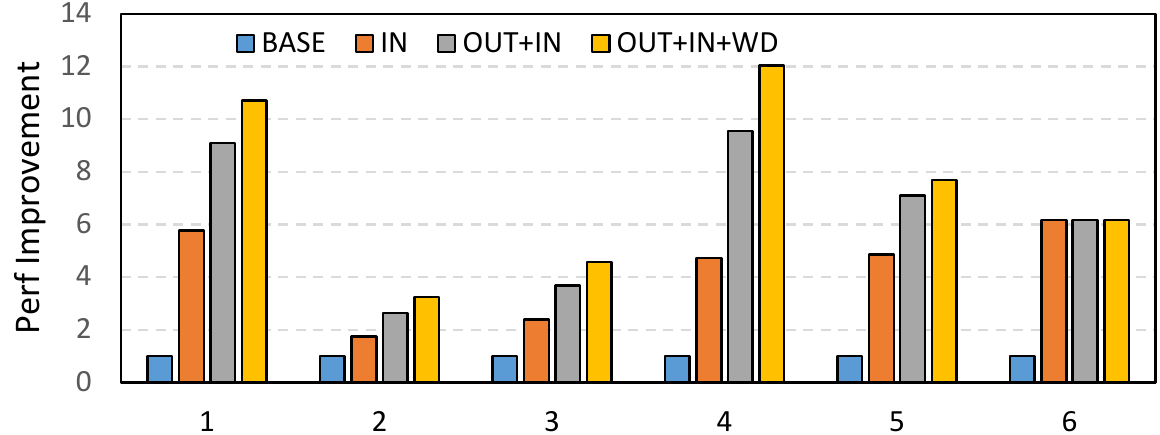}
        \caption{Inception 3b (from GoogLeNet).}
        \label{fig:inception_3a}
    \end{subfigure}%
    
    \caption{Layer-wise performance improvement results.}
    \label{fig:VGG_google}
    \vspace{-5mm}
\end{figure}



\section{\textbf{Experimental Results}} \label{exp_results}

In this section, we discuss the effects of input sparsity (IN) and output sparsity (OUT) exploitation in CNNs.
We report the overall results for both forward and backward propagation of CNN training to show the efficacy of our proposed mechanism. We evaluate four different scenarios to analyze and evaluate layer-wise performance, as shown in Fig.~\ref{fig:VGGNetLayerwise}.
In the {\em baseline scenario}, we evaluate a dense compute model (DC) and report all other results normalized with respect to the DC model.
Next, we evaluate improvements due to input sparsity, in which gradient sparsity of layer $(l+1)$ is exploited for computing the error gradient of layer $l$ (referred to as IN). 
The third bar shows the results with both the input and output sparsity exploited (IN+OUT) jointly. The last bar shows the results for the IN+OUT sparsity along with the work redistribution (IN+OUT+WR) technique in place.

We discuss our application specific results under two categories: Networks (i) with and (ii) without the BN layer. Both VGG and GoogleNet are without BN layers, and enable the joint exploitation of the input and output sparsities simultaneously.

Fig.~\ref{fig:VGGNetLayerwise} shows that, each of the VGGNet layers gets significant performance boost, ranging from  1.46$\times$ (layer 8) to 7.61$\times$ (layer 7) when employing the proposed sparsity exploitation and work redistribution schemes. Note that output sparsity exploitation is not applicable to certain layers (the 3rd, 5th, 8th and 11th bars in Fig.~\ref{fig:VGGNetLayerwise}). This is because for these particular $CONV$ layers, the immediate preceding layer is a non $ReLU$ layer (in this case, a $MaxPool$ layer). At a $MaxPool-CONV$ layer boundary, all the output gradient locations must be evaluated, which is relayed back to the preceding layer. However, our design can still take advantage of input sparsity and perform better than the dense baseline. 

\begin{figure}
\centering
\begin{subfigure}{0.45\textwidth}
  \centering
    \includegraphics[width=1.0\linewidth]{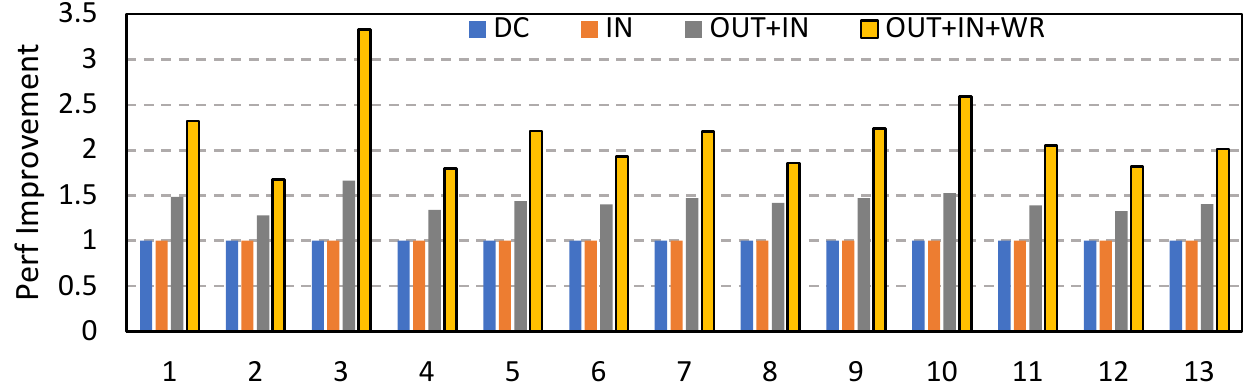}
    \caption{DenseBlock1 (DenseNet-121)}
    \label{fig:dense_block} 
\end{subfigure}%

\begin{subfigure}{0.45\textwidth}
  \centering
     \includegraphics[width=1.0\linewidth]{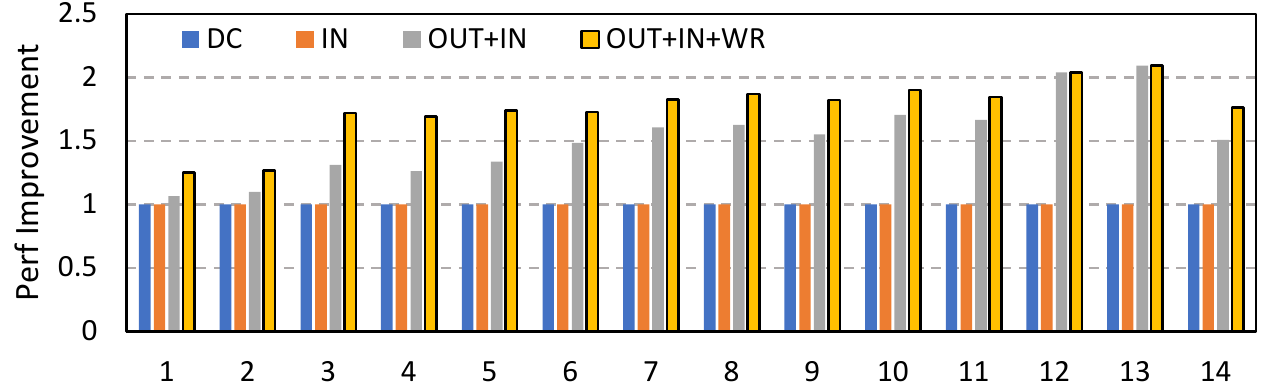}
    \caption{MobileNet}
    \label{fig:mobilenet}
\end{subfigure}%
    
    \caption{Layerwise performance improvement results.}
    \label{fig:Different Blocks}
    \vspace{-5mm}
\centering
\end{figure}

\textbf{GoogLeNet: }We only show the results for the Inception-3b module, for which the input and output sparsity exploitation can be found in  Fig.~\ref{fig:Motivation_Fig}a. he As shown, in the figure all the CONV layers within the block gets performance improvement, particularly $\{ReLU33, CONV33\}$(bar 4) and \{$ReLU55,CONV55$\}(bar 5) are direct candidates for {\em both} types of sparsity exploitation, which also happens to be the most compute-intensive CONV layers featuring 3x3 and 5x5 filter layers (others are 1x1 types). Bar-6 represents the interface between the maxpool and conv layers, and therefore, the sparsity benefits are limited to input only.
The overall achieved performance gains range from 2.6$\times$ to 12.6$\times$ for this block. 



\begin{wrapfigure}{R}{0.65\linewidth}
\vspace{-3mm}
    \centering
    \includegraphics[width=1.0\linewidth]{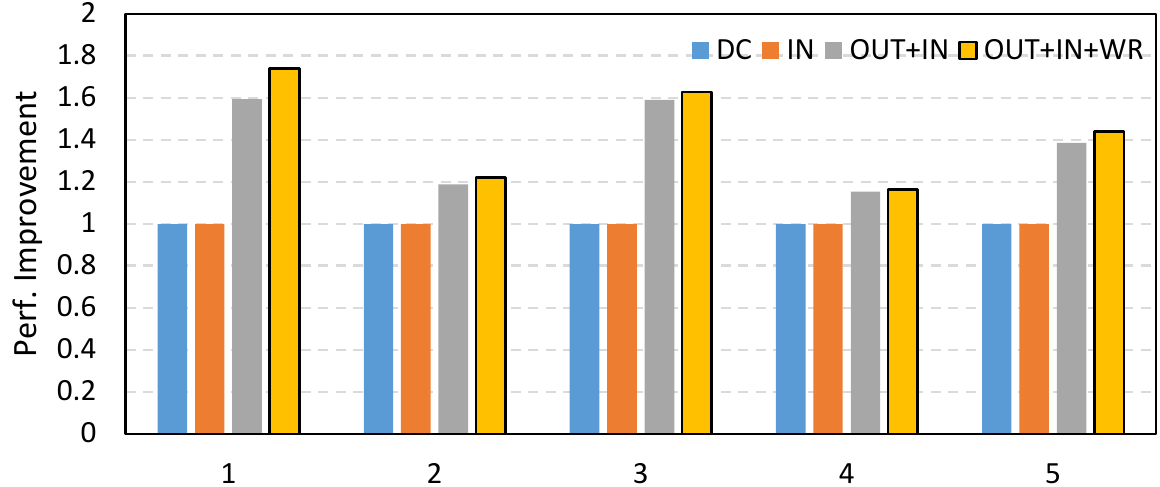}
    \caption{Residual Block2 (ResNet18)}
    \label{fig:res_block2}
\end{wrapfigure}

\textbf{Networks with the BN layer: }
For networks with the batch-normalization layers, it is important to note that the error gradients at the input of CONV layer are not sparse. Hence, the traditional input-sparsity techniques (which have been exploited by prior works) are not applicable, and only type of sparsity that is applicable is output sparsity, since these CONV layers are still preceded by ReLU layers. 
Therefore, this particular trend is observable in the results of ResNet, DenseNet and MobileNet, as shown in Figures~\ref{fig:res_block2}, \ref{fig:dense_block} and \ref{fig:mobilenet}, respectively. Both ResNet and DenseNet consist of repeating micro-blocks, and the results are shown for one such representative block.

\begin{figure}
    \centering
    \includegraphics[width=0.8\linewidth]{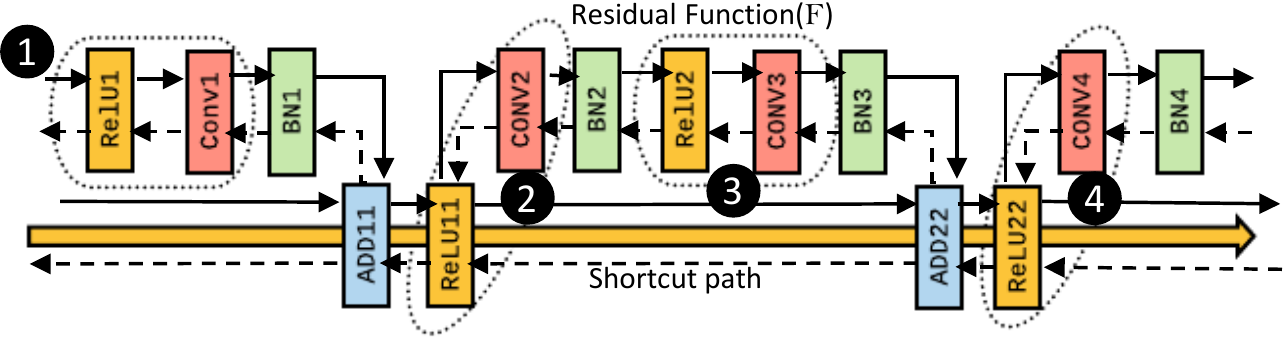}
    \caption{Residual block architecture.}
    \label{fig:res_block2_fig}
     \vspace{-5mm}
\end{figure}

\textbf{ResNet:} consists of multiple residual blocks(Fig \ref{fig:res_block2_fig}), where the output of the residual function is added to the shortcut path. This process of element-wise addition leads to reduced activation sparsity ($\sim$30\%). This explains the relatively lower gains observed for the 2nd and 4th bars (output of residual function), as compared to 1st and 3rd bars where the activation sparsity levels are higher ($\sim$50\%). Thus, the overall performance gains vary from $\sim$16\%  to  $\sim$73\% for this block (mean improvement of $\sim$45\%, ref Fig \ref{fig:res_block2}).

\textbf{DenseNet:} It has similar block-based architecture, however, the output from the residual path is merged via \textit{concatenation} instead of \textit{addition}, which retains high sparsity levels. The results corresponding to Denseblock1 of DenseNet121 are shown in Fig.~\ref{fig:dense_block}, which shows performance improvement only after the application of IN+OUT+WR and the range of improvement varies from $\sim$1.69 to 3.32$\times$ across the layers.


\textbf{MobileNet:} It has a linear structure (similar to VGGNet); however, it consists of depth-wise (dw) and point-wise (pw) convolution layers. Note that our proposed techniques are applicable to both dw and pw layer types. However, the dw layers are not a compute bottleneck, therefore we only show the results corresponding to the point-wise conv layers in Fig ~\ref{fig:mobilenet}, which shows performance improvement ranging from 1.25$\times$ to 2.1$\times$,  after applying output sparsity and work re-distribution. 



\begin{figure}[h]
    \centering
    \includegraphics[width=0.9\linewidth]{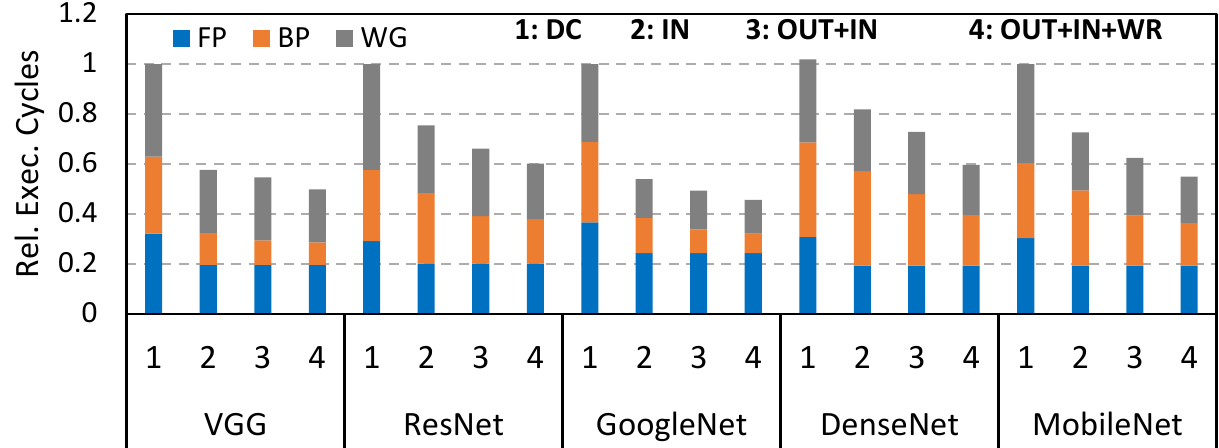}
    \caption{Normalized CNN execution time with breakdown for forward and backward passes.}
    \label{fig:EndtoENd2}
    \vspace{-4mm}
\end{figure}

Fig. ~\ref{fig:EndtoENd2} shows the overall performance improvement along with the breakdown of the contributions of the input and output sparsity exploitation techniques, taking into account both the forward and backward passes. 
Overall, the networks that benefit the most are VGGNet($\sim$ 2x) and GoogleNet($~\sim$ 2.18X) which is in line with our previous discussions as they are amenable for exploiting both input and output sparsity simultaneously in the backward pass. 
For MobileNet, ResNet and DenseNet, significant gains are still leveraged despite the presence of a BN layer, where only one type of sparsity can be exploited - input sparsity during the forward pass and output sparsity during the backward pass. The end to end benefits is 2.13x for MobileNet, 1.7x for DenseNet and 1.66x for ResNet. Note that for WG phase, OUT+IN+WR still provides improvement, as work redistribution helps in reducing the overall execution latency.

\begin{table*}[]
\begin{tabular}{|l|c|c|c|c|c|c|c|c|c|}
\hline
Platform\textbackslash{}Spec & \begin{tabular}[c]{@{}c@{}}Tech\\ (nm)\end{tabular} & \begin{tabular}[c]{@{}c@{}}Freq.\\ (MHz)\end{tabular} & \begin{tabular}[c]{@{}c@{}}Area\\ (mm2)\end{tabular} & \begin{tabular}[c]{@{}c@{}}Power\\ (W)\end{tabular} & \begin{tabular}[c]{@{}c@{}}Peak Thpt.\\ (GOps)\end{tabular} & \begin{tabular}[c]{@{}c@{}}Energy Eff. \\ (GOps/W)\end{tabular} & Exec. Mode                  & \multicolumn{2}{c|}{\begin{tabular}[c]{@{}c@{}}Iteration Latency\\   (ms)\end{tabular}} \\ \hline
                             &                                                     &                                                       &                                                      &                                                     &                                                             &                                                                 &                             & VGG-16                                     & Res-18                                     \\ \hline
Dual Xeon E5 2560 v3         & 22                                                  & 2400                                                  & -                                                    & 85                                                  & 614.4                                                       & 7.22                                                            & CPU, Dense                  & 8495                                       & 2195                                       \\ \hline
NVidia GTX 1080 Ti           & 16                                                  & 706                                                   & 400                                                  & 225                                                 & 11000                                                       & 48.8                                                            & GPU, Dense                  & 128                                        & 32.78                                      \\ \hline
DaDianNao \cite{chen2014dadiannao}                    & 65                                                  & 606                                                   & 67.3                                                 & 16.3                                                & 4964                                                        & 304                                                             & Acc, Dense                  & 526                                        & 61.1                                       \\ \hline
CNVLUTIN \cite{albericio2016cnvlutin}                 & 65                                                  & 606                                                   & 70.1                                                 & 17.4                                                & 4964                                                        & 304                                                             & Acc, Input Sparse           & 365                                        & 48.3                                       \\ \hline
LNPU \cite{LNPU}                     & 65                                                  & 200                                                   & 16                                                   & 0.367                                               & 638*                                                        & 25800*                                                          & Acc, Input Sparse           & 4742                                       & 684                                        \\ \hline
SparTANN    \cite{spartann}                 & 65                                                  & 250                                                   & 4.32                                                 & 0.59                                                & 380*                                                        & 648*                                                            & Acc, Input Sparse(BP \& WG) & 12831                                      & 1789                                       \\ \hline
Selective Grad  \cite{lee2019acceleration}             & 65                                                  & 606                                                   & 67.3                                                 & 16.3                                                & 4964                                                        & 304                                                             & Acc, Input Sparse(BP)       & 480                                        & 61.1                                       \\ \hline
This Work                    & 32                                                  & 667                                                   & 292                                                  & 19.2                                                & 5466                                                        & 325                                                             & Acc,  In + Out Sparse       & 166.81                                     & 23.26                                      \\ \hline
\end{tabular}
\caption{Comparison with CPU, GPU and prior Accelerator based platforms(* denotes throughput considering sparsity)}
\label{tab:priorwork}
\end{table*}

\begin{wrapfigure}{l}{0.5\linewidth}
    \centering
    \includegraphics[width=1.0\linewidth]{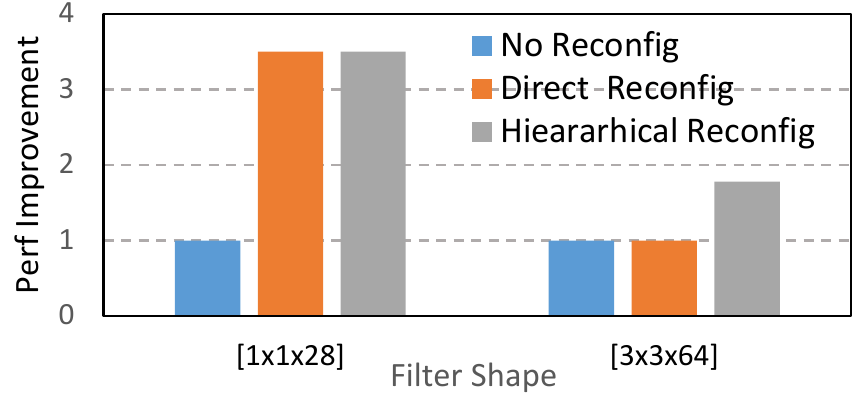}
    \caption{Impact of Lane Reconfiguration}
    \label{fig:reconfig}    
\end{wrapfigure}

\textbf{Impact of Reconfiguration:} To highlight the importance of reconfiguration, Fig. \ref{fig:reconfig} shows two types of CONV layers from Dense-block-1 from 

DenseNet, with two different receptive field sizes. The first one is [1x1x64] and occupies only 2/16 lanes, resulting in PE under-utilization. 

However, this under-utilization is readily captured in the direct form of reconfiguration, by replicating the computation 4x times. However, in the case of [3x3x64] filter size, it occupies 9/16 lanes and replication is not trivial. However, our hierarchical reconfiguration approach can schedule the computation to make up for the lost utilization otherwise, and thus it improves the performance  ($\sim$1.75x).  

\begin{wrapfigure}{l}{0.65\linewidth}
    \centering
    \includegraphics[width=1.0\linewidth]{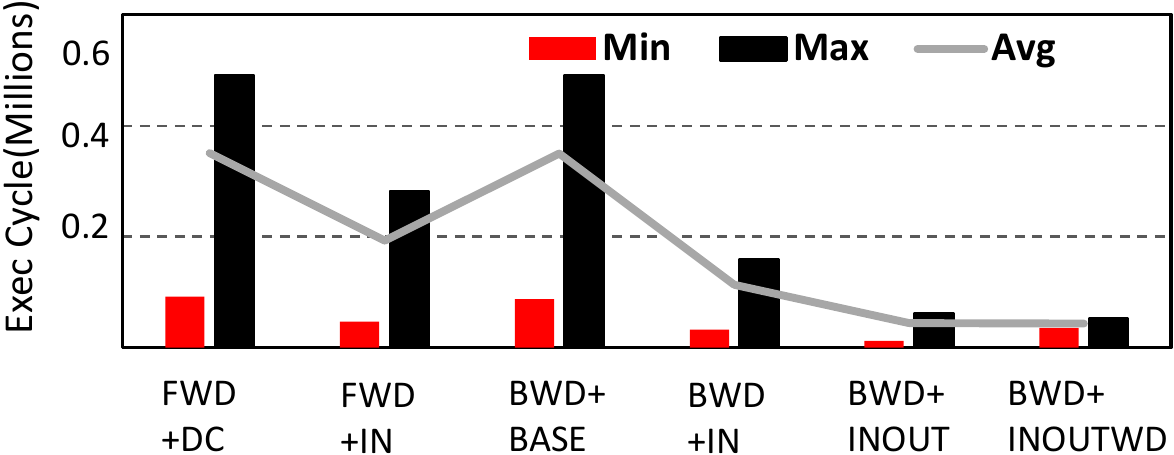}
    \caption{Node latency variation under different schemes.}
    \label{fig:utilization}
\end{wrapfigure}

\textbf{Impact on Node Utilization:} Fig.~\ref{fig:utilization} shows the minimum, maximum and average execution latencies of different tiles during GoogleNet inception 4d module execution.

As can be seen, delay is governed by the maximum tile latency which is highest for the baseline execution of both the forward and the backward cases. In general, closer the average latency curve is to the maximum latency, higher is the overall utilization and occupancy of the tile resources. We observe that, without work re-distribution, the ratio of average tile latency to maximum tile latency stays $\sim$70\%; however, with WR applied, the utilization is further improved to $\sim$82.9\%, leading to the lowest execution time and in turn savings in static energy dissipation.

\textbf{DRAM considerations:} While DRAM access pattern varies from network to network and layer to layer, for our design, we estimate that a 16 channel DDR3-1600 bandwidth (16x 12.6 GB/s) is sufficient without incurring performance loss. 
For example, for a typical layer of fmap sized at [128x28x28] and filter size of [128x128x3x3], the ratio of communication to compute time is $\sim$15\% (upto 30-50\% with sparsity). However, most of it can be overlapped with pipelined execution and hidden due to the streaming nature of data access. \textcolor{blue}{Main memory access also contributes additional $\sim$10\% (for ResNet-18) upto $\sim$ 35\% (for DenseNet-121) of total chip power.} 

\textbf {Comparison with CPUs, GPUs and prior works:}
Table \ref{tab:priorwork} shows relative design metrics of our proposed work with reference to CPUs, GPUs and prior accelerator based platforms.  We choose DaDianNao \cite{chen2014dadiannao} - a well known Dense CNN accelerator and CNVLUTIN \cite{albericio2016cnvlutin} (adapted for training), which is an input sparse variant of DaDianNao for our accelerator based comparison (both also feature identical number of MAC units (4096) and on-chip buffer size (32MB) for an apple to apple comparison with our work), in addition to three recent accelerator based designs. For CPU, we refer to  Dual Xeon E5 2630 v3 (22nm) with clock frequency of 2400 MHz, offering peak throughput of 614.4 GOps. For GPU, we refer to NVidia GTX Titan 1080 Ti (16nm) with peak 11 TOps peak throughput. Also as discussed previously, both CPU and GPU pipelines are not equipped to exploit sparsity below a high threshold, we therefore stick to the dense variant of the execution. 
The last two columns in the same table shows the performance of these different hardware as a per iteration latency(in ms) during training using batch size of 16, for VGG-16 and ResNet-18 networks. Note that CPU and GPU numbers are taken from publicly available data at \cite{cnnbench}. As seen from the table, our design perform an order of magnitude better than CPU based platform, while achieving competitive performance as compared to GPU platform. Since the GPU platform has $\sim$2x higher compute Flops and $\sim$10x more power as compared to our design, we are still able to achieve high energy efficiency($\sim$7x higher on the average) for the two benchmarks. 

Also note that, Dense variants of our proposed architecture (without sparsity specific optimizations) perform 1.9x and 1.7x better than DaDianNao, although both have similar peak throughput. This is primarily due to efficient mapping strategies leading to very high PE utilization, indicating impact of micro-architectural optimizations. Our proposed sparsity based (input and output) approach achieve 3.15x for VGG-16 and 2.65x for ResNet-18(mean 2.9x) performance improvement over DaDianNao. Over CNVLUTIN, which can only support input sparsity, improvements are in the range of 2.2x and 2.07x (mean 2.1x) over the two benchmarks.   

\textcolor{blue}{We also compare our work with three recent accelerator based proposals designed for sparse training namely LNPU \cite{LNPU}, SparTANN \cite{spartann} and Selective Grad \cite{lee2019acceleration}. LNPU is designed as a fined grained mixed precision accelerator using FP8-FP16 configurable MAC design and leverages only input sparsity. Although it reported very high energy efficiency (25.8 TFlop/W), it is considering FP8 operation at 90\% input sparsity. In addition, the design uses very limited on-chip buffer size (320 KB) as compared to our work (32MB), thus it cannot efficiently leverage the locality \& reuse properties (single weight access across input batch and input reuse across layer weights), resulting in frequent DRAM data access. This leads to an order of magnitude drop in energy efficiency at an application level. Based on our  estimates, our is work is atleast 2-3x more energy efficient compared to LNPU when considering application level characteristics, and also offers ~30x higher performance.}

\textcolor{blue}{SparTANN \cite{spartann}, on the other hand proposes to exploit threshold based sparsification for the gradient values, resulting in higher levels of sparsity. Note this work only supports input sparsity for the gradients during BP And WG stages, which is already supported in our work and doesn't discuss the scope of output sparsity and also doesn't exploit activation input sparsity during FP. Also reported energy efficiency of 648 GOps/W is considering sparse operations, which effectively translates to similar dense energy efficiency numbers as proposed in our work, while offering 14x less performance. 
Selective Grad \cite{lee2019acceleration} only exploits output sparsity of gradients in the BP, while ignoring scope of input sparsity altogether in FP, BP and WG stages. As such, our design is able to outperform \cite{lee2019acceleration} by a factor of $\sim$ 2.6x.}

\section{Conclusions}

While sparsity has been exploited in the forward \& backward passes of CNN inferences, it has been mainly in the form of \textit{input} sparsity. 
In this paper, we present a novel insight exposing the scope of output sparsity in neuron gradient computation stage, and propose a novel hardware architecture for exploiting the same, alongside available input sparsity.
To the best of our knowledge, this is the first work that jointly exploits input and output sparsity during the neuron gradient computation phase of neural network training.
Our evaluation of five state-of-the-art CNN models shows that the proposed design can reduce the execution time by up to $\sim$8.3$\times$  for the backward pass and overall by $\sim$1.81$\times$ for the CNN training step consisting of both the forward and backward passes, over a dense-compute only baseline, and achieves order of magnitude improvement in energy efficiency over dense CPU and GPU based platforms. 

\textbf{ACKNOWLEDGEMENT}

This research is supported in part by NSF grants \#1317560, \#1763681 and \#1955815.

\bibliographystyle{ieeetr}
\bibliography{ref}

\end{document}